\documentclass[12pt]{article}
\usepackage{abstract,amsmath,amssymb,latexsym}
\usepackage{enumitem,epsf}
\usepackage{fullpage,tikz,float}
\usepackage[numbers]{natbib}
\usepackage[pdftex,colorlinks]{hyperref}

\usepackage{macros}

\newif\ifdraft
\draftfalse

\title{
 \begin{minipage}[c]{1.05\textwidth}
 	\centerline{Toward Deeper Understanding of Neural Networks:}
 	\centerline{The Power of Initialization and a Dual View on Expressivity}
 \end{minipage}
}

\author{
	\vspace{1cm}
  Amit Daniely\thanks{Email: amitdaniely@google.com} \and
  Roy Frostig\thanks{Email: rf@cs.stanford.edu. Work performed at Google.} \and
  Yoram Singer\thanks{Email: singer@google.com}
}

\begin{document}

\maketitle
\thispagestyle{empty}

\begin{abstract}
We develop a general duality between neural networks and compositional kernels,
striving towards a better understanding of deep learning. We show that initial
representations generated by common random initializations are sufficiently
rich to express all functions in the dual kernel space. Hence, though the
training objective is hard to optimize in the worst case, the initial
weights form a good starting point for optimization. Our dual view also
reveals a pragmatic and aesthetic perspective of neural networks and
underscores their expressive power.

\end{abstract}

\newpage
\thispagestyle{empty}

{\large
\topskip0pt
\vspace*{\fill}
\tableofcontents
\vspace*{\fill}
}

\newpage

\setcounter{page}{1}

\section{Introduction}
Neural network (NN) learning has underpinned state of the art empirical
results in numerous applied machine learning tasks (see for
instance~\cite{krizhevsky2012imagenet,lecun2015deep}). Nonetheless, neural
network learning remains rather poorly understood in several regards.
Notably, it remains unclear why training algorithms find good weights, how
learning is impacted by the network architecture and activations, what is
the role of random weight initialization, and how to choose a concrete
optimization procedure for a given architecture.

We start by analyzing the expressive power of NNs subsequent to the random
weight initialization. The motivation is the empirical success of training
algorithms despite inherent computational intractability, and the fact that
they optimize highly non-convex objectives with potentially many local minima.
Our key result shows that random initialization already positions learning
algorithms at a good starting point. We define an object termed a {\em
computation skeleton} that describes a distilled structure of feed-forward
networks. A skeleton induces a family of network architectures along with a
hypothesis class $\ch$ of functions obtained by certain non-linear
compositions according to the skeleton's structure.  We show that the
representation generated by random initialization is sufficiently rich to
approximately express the functions in $\ch$. Concretely, all functions in
$\ch$ can be approximated by tuning the weights of the last layer, which is
a convex optimization task.

In addition to explaining in part the success in finding good weights, our
study provides an appealing perspective on neural network learning.  We
establish a tight connection between network architectures and their dual
kernel spaces. This connection generalizes several previous constructions
(see Sec~\ref{sec:related}). As we demonstrate, our dual view gives rise to
design principles for NNs, supporting current practice and suggesting
new ideas. We outline below a few points.

\begin{itemize}

\item Duals of convolutional networks appear a more suitable fit for
	vision and acoustic tasks than those of fully connected networks.

\item Our framework surfaces a principled initialization scheme. It is
	very similar to common practice, but incorporates a small correction.

\item By modifying the activation functions, two consecutive fully connected
	layers can be replaced with one while preserving the network's dual kernel.

\item The ReLU activation, i.e. $x \mapsto \max(x,0)$, possesses favorable
	properties. Its dual kernel is expressive, and it can be well approximated by
	random initialization, even when the initialization's scale is moderately
	changed.

\item As the number of layers in a fully connected network becomes very
	large, its dual kernel converges to a degenerate form for any non-linear
	activation.

\item Our result suggests that optimizing the weights of the last layer can
	serve as a convex proxy for choosing among different architectures prior
	to training. This idea was advocated and tested empirically
	in~\cite{saxe2011random}.

\end{itemize}


\section{Related work}
\label{sec:related}

\paragraph{Current theoretical understanding of NN learning.} Understanding
neural network learning, particularly its recent successes, commonly
decomposes into the following research questions.

\begin{enumerate}[label=(\roman*)]
  \item \label{related_q1} What functions can be efficiently expressed by
    neural networks?

  \item \label{related_q2} When does a low empirical loss result in a low
    population loss?

	\item \label{related_q3} Why and when do efficient algorithms, such
	as stochastic gradient, find good weights?

\end{enumerate}

\noindent Though still far from being complete, previous work provides some
understanding of questions~\ref{related_q1}~and~\ref{related_q2}. Standard
results from complexity theory~\cite{karp1980some} imply that essentially all
functions of interest (that is, any efficiently computable function) can be
expressed by a network of moderate size. Biological phenomena show that many
relevant functions can be expressed by even simpler networks, similar to
convolutional neural networks~\cite{lecun1998gradient} that are dominant in ML
tasks today. Barron's theorem~\cite{Barron93} states that even two-layer
networks can express a very rich set of functions. As for
question~\ref{related_q2}, both classical~\cite{BaumHa89, Bartlett98,
AnthonyBa99} and more recent~\cite{behnam2015norm,hardt2015train} results from
statistical learning theory show that as the number of examples grows in
comparison to the size of the network the empirical loss must be close to the
population loss. In contrast to the first two, question~\ref{related_q3} is
rather poorly understood. While learning algorithms succeed in practice,
theoretical analysis is overly pessimistic. Direct interpretation of
theoretical results suggests that when going slightly deeper beyond single
layer networks, e.g.\ to depth two networks with very few hidden units, it is
hard to predict even marginally better than random~\cite{KearnsVa89,
KlivansSh06, danielySh2014, daniely2013average, daniely2015complexity}.
Finally, we note that the recent empirical successes of NNs have prompted a
surge of theoretical work around NN learning \cite{safran2015basin,
andoni2014learning, arora2014provable, bruna2013invariant, neyshabur2015path,
livni2014computational, giryes2015deep, sedghi2014provable,
choromanska2015loss}.

\paragraph{Compositional kernels and connections to networks.} The idea of
composing kernels has repeatedly appeared throughout the machine learning
literature, for instance in early work by~\citet{scholkopf1998prior,
grauman2005pyramid}. Inspired by deep networks' success, researchers
considered deep composition of kernels~\cite{mairal2014convolutional,
cho2009kernel, bo2011object}. For fully connected two-layer networks, the
correspondence between kernels and neural networks with random weights has
been examined in~\cite{rahimi2009weighted, RahimiRe07, neal2012bayesian,
williams1997infinite}. Notably, Rahimi and Recht~\cite{rahimi2009weighted}
proved a formal connection (similar to ours) for the RBF kernel. Their work
was extended to include polynomial kernels~\cite{kar2012random,
pennington2015spherical} as well as other kernels~\cite{bach2015equivalence,
bach2014breaking}. Several authors have further explored ways to extend this
line of research to deeper, either fully-connected
networks~\cite{cho2009kernel} or convolutional networks~\cite{hazan2015steps,
anselmi2015deep, mairal2014convolutional}.  Our work sets a common foundation
for and expands on these ideas. We extend the analysis from fully-connected
and convolutional networks to a rather broad family of architectures. In
addition, we prove approximation guarantees between a network and its
corresponding kernel in our more general setting. We thus extend previous
analyses that only applies to fully connected two-layer networks. Finally, we
use the connection as an analytical tool to reason about architectural design
choices.

\section{Setting}

\paragraph{Notation.} We denote
vectors by bold-face letters (e.g.\ $\x$), and
matrices by upper case Greek letters (e.g.\ $\Sigma$). The $2$-norm of $\x
\in \reals^d$ is denoted by $\|\x\|$. For functions $\sigma:\reals\to\reals$
we let
$$
\|\sigma\| \textstyle
	:=\sqrt{\E_{X\sim\cn(0,1)}\sigma^2(X)}
	\; = \sqrt{\frac{1}{\sqrt{2\pi}}
		\int_{-\infty}^\infty \sigma^2(x)e^{-\frac{x^2}{2}}dx} \,.
$$
Let $G=(V,E)$ be a directed acyclic graph. The set of neighbors incoming to
a vertex $v$ is denoted $\IN(v):=\{u\in V\mid uv\in E\}$.
The $d-1$ dimensional sphere is denoted $\sphere^{d-1} =
\{\x\in\reals^d \mid \|\x\|=1\}$. We provide a brief overview of
reproducing kernel Hilbert spaces in the sequel and merely introduce
notation here. In a Hilbert space $\ch$, we use a slightly
non-standard notation $\ch^B$ for the ball of radius $B$, $\{\x \in
\ch \mid \|\x\|_\ch \leq B\}$. We use $[x]_+$ to denote $\max(x,0)$
and $\ind[b]$ to denote the indicator function of a binary variable
$b$.

\paragraph{Input space.} Throughout the paper we assume that each example is
a sequence of $n$ elements, each of which is represented as a unit vector. 
Namely, we fix $n$ and take the input space to be
	$\cx=\cx_{n,d}=\left(\sphere^{d-1}\right)^n$.
Each input example is denoted,
\begin{align} \label{eq:coordinates}
	\x=(\x^1,\ldots,\x^n), ~\textwhere \x^i\in \sphere^{d-1} \,.
\end{align}
We refer to each vector $\x^i$ as the input's $i$th {\em
  coordinate}, and use $x^i_{j}$ to denote it $j$th scalar
entry. Though this notation is slightly non-standard, it unifies 
input types seen in various domains. For example,
binary features can be encoded by taking $d=1$, in which case
$\cx=\{\pm 1\}^n$. Meanwhile, images and audio signals are often
represented as bounded and continuous numerical values---we can assume
in full generality that these values lie in $[-1,1]$. To match the setup
above, we embed $[-1,1]$ into the circle $\sphere^1$, e.g.\ via the map $x
\mapsto \left(\sin\left(\frac{\pi x}{2}\right), \cos\left(\frac{\pi
  x}{2}\right)\right)$. When each coordinate is categorical---taking
one of $d$ values---we can represent category $j\in[d]$ by the unit
vector $\mathbf e_j\in\sphere^{d-1}$. When $d$ may be very large or
the basic units exhibits some structure, such as when the input is a sequence
of words, a more concise encoding may be
useful, e.g.\ as unit vectors in a low dimension space $\sphere^{d'}$
where $d'\ll d$ (see for
instance~\citet{mikolov2013distributed,levy2014neural}).

\paragraph{Supervised learning.} The goal in supervised learning is to
devise a mapping from the input space $\cx$ to an output space $\cy$ based on a sample
$S=\{(\x_1,y_1),\ldots,(\x_m,y_m)\}$, where $(\x_i,y_i)\in\cx\times\cy$, 
drawn i.i.d.\ from a distribution $\cd$ over $\cx\times\cy$.
A supervised learning problem is further specified by an output length $k$ and a loss function
$\ell : \reals^k \times \cy \to [0,\infty)$, and the goal is to find a
predictor $h:\cx\to\reals^k$ whose loss,
$\cl_{\cd}(h) := \E_{(\x,y)\sim\cd} \ell(h(\x),y)$, is small.
The {\em empirical} loss $\cl_{S}(h):= \frac 1 m \sum_{i=1}^m
\ell(h(\x_i),y_i)$ is commonly used as a proxy for the loss
$\cl_{\cd}$.
Regression problems correspond to $\cy=\reals$ and, for
instance, the squared loss $\ell(\hat y,y)=(\hat y -y)^2$.
Binary classification is captured by $\cy=\{\pm 1\}$ and, say, the
zero-one loss $\ell(\hat y,y)= \ind[\hat y y \leq 0]$ or the hinge
loss $\ell(\hat y,y)=[1-\hat y y]_+$, with standard extensions to the
multiclass case.
A loss $\ell$ is $L$-Lipschitz if $|\ell(y_1,y) - \ell(y_2,y) | \leq L
|y_1 - y_2|$ for all $y_1,y_2 \in \reals^k$, $y \in \cy$, and it is
convex if $\ell(\cdot,y)$ is convex for every $y\in\cy$.

\paragraph{Neural network learning.} We define a {\em neural network} $\cn$
to be a vertices weighted directed acyclic graph (DAG) whose nodes are denoted $V(\cn)$ and edges
$E(\cn)$. The weight function will be denoted by $\delta:V(\cn)\to [0,\infty)$, and its sole role would be to dictate the distribution of the initial weights (see definition \ref{def:rand_weights}).
Each of its internal units, i.e.\ nodes with both incoming and
outgoing edges, is associated with an {\em activation} function
$\sigma_v:\reals\to\reals$. In this paper's context, an activation can
be any function that is square integrable with respect to the Gaussian
measure on $\reals$. We say that $\sigma$ is {\em normalized} if
$\|\sigma\|=1$. The set of nodes having only incoming edges are called the
output nodes.
To match the setup of a supervised learning problem, a network $\cn$ has
$nd$ input nodes and $k$ output nodes, denoted $o_1,\ldots,o_k$. A
network $\cn$ together with a weight vector $\w=\{w_{uv} \mid uv\in E\}$ defines a
predictor $h_{\cn,\w}:\cx\to\reals^k$ whose prediction
is given by ``propagating'' $\x$ forward through the network.
Formally, we define $h_{v,\w}(\cdot)$ to be the output of the subgraph
of the node $v$ as follows: for an input node $v$, $h_{v,\w}$ outputs the corresponding coordinate in $\x$, and
for all other nodes, we define $h_{v,\w}$ recursively as
$$h_{v,\w}(\x) = \sigma_v\left(\textstyle
	\sum_{u\in \IN(v)}\, w_{uv}\,h_{u,\w}(\x)\right)\,.$$
Finally, we let $h_{\cn,\w}(\x)=(h_{o_1,\w}(\x),\ldots,h_{o_k,\w}(\x))$.
We also refer to internal nodes as {\em hidden units}. The {\em output
layer} of $\cn$ is the sub-network consisting of all output neurons of $\cn$
along with their incoming edges. The {\em representation} induced by a network
$\cn$ is the network $\netrep(\cn)$ obtained from $\cn$ by removing the output
layer. The {\em representation} function induced by the weights $\w$ is
$\rep_{\cn,\w}:=h_{\netrep(\cn),\w}.$
Given a sample $S$, a learning algorithm searches
for weights $\w$ having small empirical loss
$\cl_S(\w)=\frac{1}{m}\sum_{i=1}^m \ell(h_{\cn,\w}(\x_i),y_i)$. A popular
approach is to randomly initialize the weights and then use a variant
of the stochastic gradient method to improve these weights in the
direction of lower empirical loss.

\paragraph{Kernel learning.} A function $\kappa:\cx\times \cx\to \reals$ is
a {\em reproducing kernel}, or simply a kernel, if for every
$\x_1,\ldots,\x_r\in\cx$, the $r \by r$ matrix $\Gamma_{i,j} = \{\kappa(\x_i,\x_j)\}$
is positive semi-definite. Each kernel induces a Hilbert space
$\ch_{\kappa}$ of functions from $\cx$ to $\reals$ with a corresponding norm
$\|\cdot\|_{\ch_\kappa}$. A kernel and its corresponding space are {\em
normalized} if $\forall \x\in\cx,\;\kappa(\x,\x)=1$. Given a convex loss
function $\ell$, 
a sample $S$, and a kernel
$\kappa$, a kernel learning algorithm finds a function
$f=(f_1,\ldots,f_k)\in\ch^k_{\kappa}$ whose empirical loss,
$\cl_S(f)=\frac{1}{m}\sum_i \ell(f(\x_i),y_i)$, is minimal among all
functions with $\sum_{i}\|f_i\|_\kappa^2 \le R^2$ for some $R>0$.
Alternatively, kernel algorithms minimize the {\em regularized loss},
$$\cl^R_S(f)=\frac{1}{m}\sum_{i=1}^m \ell(f(\x_i),y_i) +
\frac{1}{R^2}\sum_{i=1}^k\|f_i\|^2_{\kappa} \,, $$
a convex objective that often can be efficiently minimized.

\section{Computation skeletons}
In this section we define a simple structure which we term a computation
skeleton. The purpose of a computational skeleton is to compactly describe
a feed-forward computation from an input to an output. A single skeleton encompasses a family of neural networks
that share the same skeletal structure. Likewise, it defines a
corresponding kernel space.
\begin{definition} A {\em computation skeleton} $\cs$ is a DAG whose
non-input nodes are labeled by activations.
\end{definition}
Though the formal definition of neural networks and skeletons appear
identical, we make a conceptual distinction between them as their role in
our analysis is rather different. Accompanied by a set of weights, a neural
network describes a concrete function, whereas the skeleton stands for a
topology common to several networks as well as for a kernel.
To further underscore the differences we note that skeletons are naturally more compact than networks. In particular, all examples of
skeletons in this paper are {\em irreducible}, meaning that for
each two nodes $v,u\in{}V(\cs)$, $\IN(v)\neq\IN(u)$.
We further restrict our attention
to skeletons with a single output node, showing later that single-output
skeletons can capture supervised problems with outputs in $\reals^k$. We
denote by $|\cs|$ the number of non-input nodes of $\cs$.

\begin{figure}[t]
\begin{center}
\begin{tikzpicture}
\foreach \i in {1,...,4}
{
	\draw (\i -7,0) rectangle (\i-7+0.5,0.5);
	\draw [->] (\i-7 +0.25  ,0.5) -- (-4.25 ,1);
}
\draw (-4.5,1) rectangle (-4,1.5);
\draw [->] (-4.25 ,1.5) -- (-4.25 ,2);
\node[text width=3cm] at (-2.85,-0.25) {$\cs_1$};
\foreach \i in {1,...,4}
{
	\draw (\i -1,0) rectangle (\i-1+0.5,0.5);
}
\foreach \i in {1,...,3}
{
	\draw (\i-1 + 0.5 ,1) rectangle (\i-1+1,1.5);
	\draw [->] (\i-1 +0.25  ,0.5) -- (\i-1 + 0.75 ,1);
	\draw [->] (\i-1 +1.25  ,0.5) -- (\i-1 + 0.75 ,1);
	\draw [->] (\i-1 + 0.75 ,1.5) -- (1.75 ,2);
}
\draw (1.5 ,2) rectangle (2,2.5);
\draw [->] (1.75 ,2.5) -- (1.75 ,3);
\node[text width=3cm] at (3.15,-0.25) {$\cs_2$};
\end{tikzpicture}
\begin{tikzpicture}
\foreach \i in {1,...,4}
{
	\draw (\i -7,0) rectangle (\i-7+0.5,0.5);
}
\foreach \i in {1,...,3}
{
	\draw (\i-7 + 0.5 ,1) rectangle (\i-7+1,1.5);
	\draw [->] (\i-7 +0.25  ,0.5) -- (\i-7 + 0.75 ,1);
	\draw [->] (\i-7 +1.25  ,0.5) -- (\i-7 + 0.75 ,1);
	\draw [->] (\i-7 + 0.75 ,1.5) -- (-2.75 ,2);
}
\draw (-3 ,2) rectangle (-2.5,2.5);
\foreach \i in {1,...,3}
{
	\draw (\i-7 + 0.5 ,3) rectangle (\i-7+1,3.5);
	\draw [->] (\i-7 +0.75, 1.5) -- (\i-7 +0.75,3);
	\draw [->] (-2.75, 2.5) -- (\i-7 +0.75,3);
	\draw [->] (\i-7 + 0.75 ,3.5) -- (-4.25 ,4);
}
\draw (-4.5 ,4) rectangle (-4,4.5);
\draw [->] (-4.25 ,4.5) -- (-4.25 ,5);
\node[text width=3cm] at (-2.85,-0.25) {$\cs_3$};
\foreach \i in {1,...,4}
{
	\draw (\i -1,0) rectangle (\i-1+0.5,0.5);
}
\foreach \i in {1,...,3}
{
	\draw (\i-1 + 0.5 ,1) rectangle (\i-1+1,1.5);
	\draw [->] (\i-1 +0.25  ,0.5) -- (\i-1 + 0.75 ,1);
	\draw [->] (\i-1 +1.25  ,0.5) -- (\i-1 + 0.75 ,1);
	\draw [->] (\i-1 + 0.75 ,1.5) -- (\i-1 + 0.75 ,2);
}
\draw [->] (1 ,1.25) -- (1.5 ,1.25);
\draw [->] (2 ,1.25) -- (2.5 ,1.25);
\foreach \i in {1,...,3}
{
	\draw (\i-1 + 0.5 ,2) rectangle (\i-1+1,2.5);
	\draw [->] (\i-1 + 0.75 ,2.5) -- (\i-1 + 0.75 ,3);
}
\draw [->] (1.5 ,2.25) -- (1 ,2.25);
\draw [->] (2.5 ,2.25) -- (2 ,2.25);
\node[text width=3cm] at (3.15,-0.25) {$\cs_4$};
\end{tikzpicture}
\caption{Examples of computation skeletons.\label{fig:cs_examples}}
\end{center}
\end{figure}
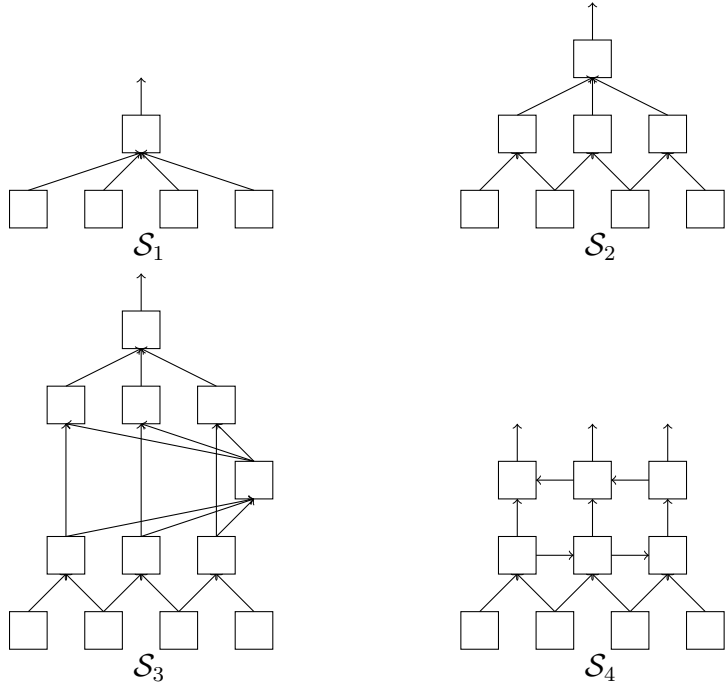

Figure \ref{fig:cs_examples} shows four example skeletons, omitting
the designation of the activation functions. The skeleton
$\cs_1$ is rather basic as it aggregates all the inputs in a single
step. Such topology can be useful in the absence of any prior
knowledge of how the output label may be computed from an input example, and
it is commonly used in natural language processing where the input is
represented as a bag-of-words~\cite{harris1954distributional}. The only structure in $\cs_1$
is a single {\em fully connected} layer:
\begin{terminology}[Fully connected layer of a skeleton]
An induced subgraph of a skeleton with $r+1$ nodes, $u_1,\ldots,u_r,v$, is
called a {\em fully connected layer} if its edges are
$u_1v,\ldots,u_rv$.
\end{terminology}
The skeleton $\cs_2$ is slightly more involved: it first processes
consecutive (overlapping) parts of the input, and the next layer
aggregates the partial results. Altogether, it corresponds to networks
with a single one-dimensional convolutional layer, followed by a fully
connected layer. The two-dimensional (and deeper) counterparts of such skeletons
correspond to networks that are common in visual object recognition.
\begin{terminology}[Convolution layer of a skeleton]
Let $s,w,q$ be positive integers and denote $n=s(q-1)+w$. A subgraph
of a skeleton is a one dimensional {\em convolution layer} of width $w$ and stride $s$
if it has $n+q$ nodes, $u_1,\ldots,u_{n}, v_1,\ldots,v_q$, and $q w$ edges,
$u_{s(i-1)+j}\,v_{i}$, for $1\le i\le q,1\le j\le w$.
\end{terminology}
The skeleton $\cs_3$ is a somewhat more sophisticated version of $\cs_2$: the local computations are first aggregated, then reconsidered with the aggregate, and finally aggregated again.
The last skeleton, $\cs_4$, corresponds to the networks that arise in
learning sequence-to-sequence mappings as used in translation, speech
recognition, and OCR tasks (see for example~\citet{sutskever2014sequence}).

\subsection{From computation skeletons to neural networks}
%
The following definition shows how a skeleton, accompanied with a
replication parameter $r\ge 1$ and a number of output nodes $k$,
induces a neural network architecture. Recall that inputs are ordered
sets of vectors in $\sphere^{d-1}$.
%
\begin{definition}[Realization of a skeleton]
Let $\cs$ be a computation skeleton and consider input coordinates in
$\sphere^{d-1}$ as in \eqref{eq:coordinates}. For $r, k \ge 1$ we
define the following neural network $\cn=\cn(\cs,r,k)$.
For each input node in $\cs$, $\cn$ has $d$ corresponding input
neurons with weight $1/d$. For each internal node $v\in \cs$ labeled by an activation
$\sigma$, $\cn$ has $r$ neurons $v^1,\ldots,v^r$, each with an
activation $\sigma$ and weight $1/r$. In addition, $\cn$ has $k$ output neurons
$o_1,\ldots,o_k$ with the identity activation $\sigma(x)=x$ and weight $1$.
There is an edge $v^iu^j\in E(\cn)$ whenever $uv\in
E(\cs)$.  For every output node $v$ in $\cs$, each neuron $v^j$ is
connected to all output neurons $o_1,\ldots,o_k$. We term $\cn$ the
{\em $(r,k)$-fold realization} of $\cs$. We also define the {\em $r$-fold realization} of $\cs$ as\footnote{Note that for every $k$,
$\netrep\left(\cn(\cs,r,1)\right)=\netrep\left(\cn(\cs,r,k)\right)$.} $\cn(\cs,r)= \netrep\left(\cn(\cs,r,1)\right)$.
\end{definition}

\noindent
Note that the notion of the replication parameter $r$ corresponds, in
the terminology of convolutional networks, to the number of
channels taken in a convolutional layer and to the number of hidden units
taken in a fully-connected layer.

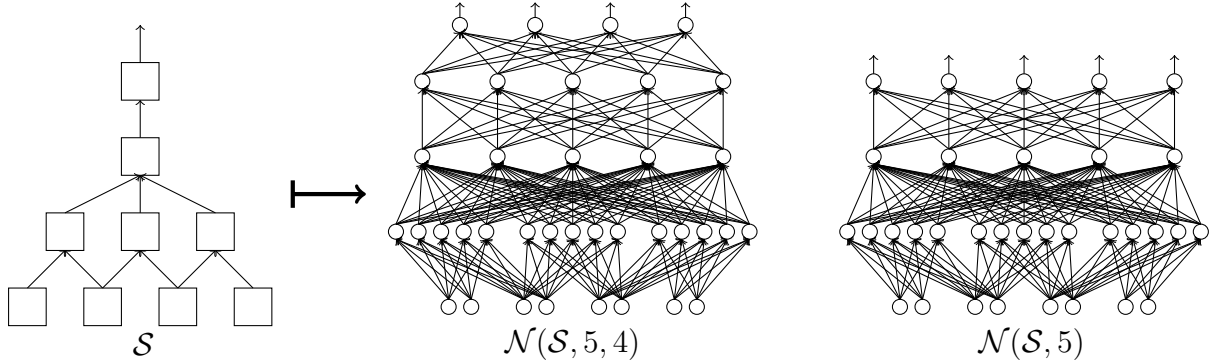
\begin{figure}[t]
\begin{tikzpicture}
\foreach \i in {1,...,4}
{
	\draw (\i -1,0) rectangle (\i-1+0.5,0.5);
}
\foreach \i in {1,...,3}
{
	\draw (\i-1 + 0.5 ,1) rectangle (\i-1+1,1.5);
	\draw [->] (\i-1 +0.25  ,0.5) -- (\i-1 + 0.75 ,1);
	\draw [->] (\i-1 +1.25  ,0.5) -- (\i-1 + 0.75 ,1);
	\draw [->] (\i-1 + 0.75 ,1.5) -- (1.75 ,2);
}
\draw (1.5 ,2) rectangle (2,2.5);
\draw [->] (1.75 ,2.5) -- (1.75 ,3);
\draw (1.5 ,3) rectangle (2,3.5);
\draw [->] (1.75 ,3.5) -- (1.75 ,4);
\node[text width=3cm] at (3.15,-0.25) {$\cs$};
\draw [ultra thick, |->] (3.75,1.75) -- (4.75,1.75);
\foreach \i in {1,...,4}
{
	\draw (4.85+\i,0.25) circle [radius=0.1];
	\draw (5.15+\i,0.25) circle [radius=0.1];
}
\foreach \i in {-1,...,1}
{
	\foreach \j in {-2,...,2}
	{
		\draw (7.5+1.75*\i + 0.3*\j,1.25) circle [radius=0.1];
		\draw [->] (6.85+\i,0.35) -- (7.5+1.75*\i + 0.3*\j,1.15);
		\draw [->] (7.15+\i,0.35) -- (7.5+1.75*\i + 0.3*\j,1.15);

		\draw [->] (7.85+\i,0.35) -- (7.5+1.75*\i + 0.3*\j,1.15);
		\draw [->] (8.15+\i,0.35) -- (7.5+1.75*\i + 0.3*\j,1.15);
		\foreach \k in {-2,...,2}
		{
			\draw [->] (7.5+1.75*\i + 0.3*\j,1.35) -- (7.5+\k,2.15);
		}
	}
}
\foreach \j in {-2,...,2}
{
	\draw (7.5+\j,2.25) circle [radius=0.1];
	\foreach \k in {-2,...,2}
	{
		\draw[->] (7.5+\j,2.35) -- (7.5+\k,3.15);
	}
}
\foreach \j in {-2,...,2}
{
	\draw (7.5+\j,3.25) circle [radius=0.1];
	\draw[->] (7.5+\j,3.35) -- (6,3.9);
	\draw[->] (7.5+\j,3.35) -- (7,3.9);
	\draw[->] (7.5+\j,3.35) -- (8,3.9);
	\draw[->] (7.5+\j,3.35) -- (9,3.9);
}
\draw (6,4) circle [radius=0.1];
\draw (7,4) circle [radius=0.1];
\draw (8,4) circle [radius=0.1];
\draw (9,4) circle [radius=0.1];
\draw[->] (6,4.1) -- (6,4.3);
\draw[->] (7,4.1) -- (7,4.3);
\draw[->] (8,4.1) -- (8,4.3);
\draw[->] (9,4.1) -- (9,4.3);
\node[text width=3cm] at (8.1,-0.25) {$\cn(\cs,5,4)$};
\foreach \i in {1,...,4}
{
	\draw (10.85+\i,0.25) circle [radius=0.1];
	\draw (11.15+\i,0.25) circle [radius=0.1];
}
\foreach \i in {-1,...,1}
{
	\foreach \j in {-2,...,2}
	{
		\draw (6+7.5+1.75*\i + 0.3*\j,1.25) circle [radius=0.1];
		\draw [->] (6+6.85+\i,0.35) -- (6+7.5+1.75*\i + 0.3*\j,1.15);
		\draw [->] (6+7.15+\i,0.35) -- (6+7.5+1.75*\i + 0.3*\j,1.15);

		\draw [->] (6+7.85+\i,0.35) -- (6+7.5+1.75*\i + 0.3*\j,1.15);
		\draw [->] (6+8.15+\i,0.35) -- (6+7.5+1.75*\i + 0.3*\j,1.15);
		\foreach \k in {-2,...,2}
		{
			\draw [->] (6+7.5+1.75*\i + 0.3*\j,1.35) -- (6+7.5+\k,2.15);
		}
	}
}
\foreach \j in {-2,...,2}
{
	\draw (6+7.5+\j,2.25) circle [radius=0.1];
	\foreach \k in {-2,...,2}
	{
		\draw[->] (6+7.5+\j,2.35) -- (6+7.5+\k,3.15);
	}
}
\foreach \j in {-2,...,2}
{
	\draw (6+7.5+\j,3.25) circle [radius=0.1];
	\draw[->] (6+7.5+\j,3.35) -- (6+7.5+\j,3.6);
}
\node[text width=3cm] at (6+8.4,-0.25) {$\cn(\cs,5)$};
\end{tikzpicture}
\caption{A $(5,4)$-fold and $5$-fold realizations of the computation
skeleton $\cs$ with $d=2$.\label{fig:ct_to_nn}}
\end{figure}

Figure~\ref{fig:ct_to_nn} illustrates a $(5,4)$- and $5$-realizations
of a skeleton with coordinate dimension $d=2$.  The
$(5,4)$-realization is a network with a single (one dimensional)
convolutional layer having $5$ channels, stride of $2$, and width of
$4$, followed by three fully-connected layers. The global replication
parameter $r$ in a realization is used for brevity; it is
straightforward to extend results when the different nodes in $\cs$
are each replicated to a different extent.


We next define a scheme for random initialization of the weights of a neural
network, that is similar to what is often done in practice. We employ the definition throughout the paper whenever we refer to
random weights.
\begin{definition}[Random weights]\label{def:rand_weights}
A {\em random initialization} of a neural network $\cn$ is a
multivariate Gaussian $\w=(w_{uv})_{uv\in E(\cn)}$ such that each weight
$w_{uv}$ is sampled independently from a normal distribution with mean $0$
and variance\footnote{For $U\subset V(\cn)$ we denote $\delta(U) = \sum_{u\in U}\delta(u)$.} ${d\delta(u)}/{\delta(\IN(v))}$ if $u$ is an input neuron and ${\delta(u)}/{\left(\|\sigma_{u}\|^2\,\delta(\IN(v))\right)}$ otherwise.
\end{definition}
%
\noindent
Architectures such as convolutional nets have weights that are shared
across different edges.
Again, it is straightforward to extend our results to these cases and for
simplicity we assume no explicit weight sharing.

\subsection{From computation skeletons to reproducing kernels}
In addition to networks' architectures, a computation skeleton $\cs$ also
defines a normalized kernel $\kappa_\cs:\cx\times\cx\to[-1,1]$ and a
corresponding norm $\|\cdot \|_{\cs}$ on functions $f:\cx\to\reals$. This
norm has the property that $\|f\|_{\cs}$ is small if and only if $f$ can be
obtained by certain simple compositions of functions according to the
structure of $\cs$. To define the kernel,
we introduce a {\em dual activation} and {\em dual kernel}. For
$\rho\in[-1,1]$, we denote by $\gaussian_\rho$ the multivariate Gaussian
distribution on $\reals^2$ with mean $0$ and covariance matrix
$\left( \begin{smallmatrix} 1 & \rho \\ \rho & 1 \end{smallmatrix} \right)$.
%

\begin{definition}[Dual activation and kernel]\label{def:dual_act}
The {\em dual activation} of an activation $\sigma$ is the function
$\hat{\sigma}:[-1,1]\to\reals$ defined as
$$
\hat\sigma(\rho) =
	\E_{(X,Y) \sim \gaussian_\rho}\sigma(X)\sigma(Y) \,.
$$
The {\em dual kernel} w.r.t.\ to a Hilbert space $\ch$ is the kernel
$\kappa_\sigma:\ch^1\times \ch^1\to\reals$ defined as
$$\kappa_\sigma(\x,\y) = \hat{\sigma}(\inner{\x,\y}_\ch) \,.$$

\end{definition}
\noindent
Section~\ref{sec:comp_ker} shows that $\kappa_\sigma$ is indeed a kernel for every activation
$\sigma$ that adheres with the square-integrability requirement.
In fact,
any continuous $\mu:[-1,1]\to\reals$, such that
$(\x,\y)\mapsto\mu(\inner{\x,\y}_\ch)$ is a kernel for all $\ch$, is
the dual of some activation.
Note that $\kappa_\sigma$ is normalized iff $\sigma$ is normalized.
We show in Section~\ref{dualact:sec} that dual activations are closely
related to Hermite polynomial expansions, and that these can be used
to calculate the duals of activation functions analytically.
Table~\ref{tab:duals} lists a few examples of normalized activations
and their corresponding dual (corresponding derivations are in
Section~\ref{dualact:sec}).
{\renewcommand{\arraystretch}{1.3}%
\begin{table}[t]
\begin{center}
  \begin{tabular}{lllll}
    \hline
    Activation &  & Dual Activation & Kernel & Ref \\ \hline
    Identity & $x$ & $\rho$ & linear\\
    2nd Hermite & $\frac{x^2 - 1}{\sqrt{2}}$ & $\rho^2$ & poly &\\
    ReLU & $\sqrt{2}\,[x]_+$ &
		$\frac{1}{\pi}+\frac{\rho}{2} +
		 \frac{\rho^2}{2\pi} +
		 \frac{\rho^4}{24\pi} + \ldots
		 = \frac{\sqrt{1-\rho^2}+(\pi-\cos^{-1}(\rho))\rho}{\pi}$ &
		 $\arccos_1$  & \cite{cho2009kernel}\\
    Step & $\sqrt{2}\,\ind[{x\ge 0}]$ &
			$\frac{1}{2} +
			 \frac{\rho}{\pi} +
			 \frac{\rho^3}{6\pi} +
			 \frac{3\rho^5}{40\pi} + \ldots = \frac{\pi-\cos^{-1}(\rho)}{\pi}$ &
			 $\arccos_0$ & \cite{cho2009kernel}\\
    Exponential & $e^{x-2}$ &
			$\frac{1}{e}+\frac{\rho}{e}+\frac{\rho^2}{2e}+\frac{\rho^3}{6e}+\ldots=
			e^{\rho-1}$ & RBF & \cite{mairal2014convolutional}\\
			& & & & \vspace{-16pt} \\
			\hline
  \end{tabular}
\caption{Activation functions and their duals.\label{tab:duals}}
\end{center}
\end{table}
}
The following definition gives the kernel corresponding to a skeleton
having normalized activations.\footnote{For a skeleton $\cs$ with unnormalized
  activations, the corresponding kernel is the kernel of the skeleton
  $\cs'$ obtained by normalizing the activations of $\cs$.}
\begin{definition}[Compositional kernels]\label{def:comp_ker}
Let $\cs$ be a computation skeleton with normalized activations and
(single) output node $o$.
For every node $v$, inductively define a kernel
$\kappa_v:\cx\times\cx\to\reals$ as follows.
For an input node $v$ corresponding to the $i$th coordinate,
define $\kappa_{v}(\x,\y)=\inner{\x^i, \y^i}$.
For a non-input node $v$, define
$$
\kappa_v(\x,\y) =
	\hat\sigma_v\left(
		\frac{\sum_{u\in \IN(v)}\kappa_{u}(\x,\y)}{|\IN(v)|}\right) \,.
    $$
The final kernel $\kappa_\cs$ is $\kappa_o$, the kernel associated with
the output node $o$. The resulting Hilbert space and norm are
denoted $\ch_\cs$ and $\|\cdot\|_\cs$ respectively, and
$\ch_v$ and $\|\cdot\|_v$ denote the space and norm when formed at node $v$.
\end{definition}
\noindent As we show later, $\kappa_\cs$ is indeed a (normalized) kernel for every
skeleton $\cs$. To understand the kernel in the context of learning, we need
to examine which functions can be expressed as moderate
norm functions in $\ch_\cs$. As we show in section \ref{sec:comp_ker},
these are the functions obtained by certain simple compositions according to the
feed-forward structure of $\cs$. For intuition, the following example
contrasts two commonly used skeletons.

\begin{example}[Convolutional vs.\ fully connected skeletons]
\label{exam:diff_struct}
Consider a network whose activations are all ReLU, $\sigma(z)=[z]_+$,
and an input space $\cx_{n,1}=\{\pm 1\}^n$. Say that $\cs_1$ is a
skeleton comprising a single fully connected layer, and that $\cs_2$ is one comprising
a convolutional layer of stride $1$ and width $q=\log^{0.999}(n)$, followed by a single fully-connected layer. (The skeleton $\cs_2$ from
Figure~\ref{fig:cs_examples} is a concrete example of the convolutional
skeleton with $q=2$ and $n=4$.)
The kernel $\kappa_{\cs_1}$ takes the form $\kappa_{\cs_1}(\x,\y) =
\hat{\sigma} \left({\inner{\x,\y}}/{n}\right)$. It is a symmetric kernel and
therefore functions with small norm in $\ch_{\cs_1}$ are essentially
low-degree polynomials. For instance, fix a bound $R=n^{1.001}$ on the norm
of the functions.
In this case, the space $\ch^{R}_{\cs_1}$ contains
multiplication of one or two input coordinates. However, multiplication of
$3$ or more coordinates are no-longer in $\ch^{R}_{\cs_1}$. Moreover, this
property holds true regardless of the choice of activation
function. On the other hand, $\ch^{R}_{\cs_2}$ contains functions
whose dependence on adjacent input coordinates is far more complex. It includes,
for instance, any function $f:\cx\to \{\pm 1\}$ that is symmetric
(i.e.\ $f(x)=f(-x)$) and that depends on $q$ adjacent coordinates
$\x_{i},\ldots,\x_{i+q}$. Furthermore, any sum of $n$ such functions is
also in $\ch^{R}_{\cs_2}$.
\end{example}

\section{Main results} \label{results:sec}
We review our main results. Let us fix a compositional kernel $\cs$. There are a
few upshots to underscore upfront. First, our analysis implies that a
representation generated by a random initialization of $\cn=\cn(\cs,r,k)$
approximates the kernel $\kappa_\cs$. The sense in which the result holds is
twofold. First, with the proper rescaling we show that
$\inner{\rep_{\cn,\w}(\x),\rep_{\cn,\w}(\x')}\approx \kappa_{\cs}(\x,\x')$.
Then, we also show that the functions obtained by composing
bounded linear functions with $\rep_{\cn,\w}$ are approximately the bounded-norm functions in $\ch_\cs$. In other words, the functions
expressed by $\cn$ under varying the weights of the last layer are
approximately bounded-norm functions in $\ch_\cs$. For simplicity, we restrict
the analysis to the case $k=1$. We also confine the analysis to either bounded
activations, with bounded first and second derivatives, or the ReLU activation.
Extending the results to a broader family of activations is left for future
work. Through this and remaining sections we use $\gtrsim$ to hide universal constants.
\begin{definition}
An activation $\sigma:\reals\to\reals$ is {\em $C$-bounded} if it is twice
continuously differentiable and $\|\sigma\|_{\infty},
\|\sigma'\|_{\infty},\|\sigma''\|_\infty\le \|\sigma\|C$.
\end{definition}
\noindent Note that many activations are $C$-bounded for some constant $C>0$. In
particular, most of the popular sigmoid-like functions such as
${1}/({1+e^{-x}})$, $\erf(x)$, ${x}/{\sqrt{1+x^2}}$, $\tanh(x)$,
and $\tan^{-1}(x)$ satisfy the boundedness requirements. We next introduce
terminology that parallels the representation layer of $\cn$ with a kernel
space. Concretely, let $\cn$ be a network whose representation part has $q$
output neurons. Given weights $\w$, the {\em normalized representation}
$\Psi_\w$ is obtained from the representation $R_{\cn,\w}$ by
dividing each output neuron $v$ by $\|\sigma_v\|\sqrt{q}$. The {\em empirical
kernel} corresponding to $\w$ is defined as
$\kappa_\w(\x,\x')=\inner{\Psi_{\w}(\x),\Psi_{\w}(\x')}$. We also define the
{\em empirical kernel space} corresponding to $\w$ as
$\ch_\w=\ch_{\kappa_\w}$. Concretely,
\[
\ch_\w=\left\{h_{\bv}(\x)=\inner{\bv,\Psi_\w(x)}\mid \bv\in\reals^q\right\}~,
\]
and the norm of $\ch_\w$ is defined as $\|h\|_\w=\inf\{\|\bv\|\mid{}h=h_\bv\}$. Our first result shows that the empirical kernel approximates
the kernel $k_\cs$.
\begin{theorem}\label{thm:main_ker}
Let $\cs$ be a skeleton with $C$-bounded activations. Let $\w$ be a random
initialization of $\cn=\cn(\cs,r)$ with
$$r \ge \frac
	{(4C^4)^{\depth(\cs)+1} \log\left({8|\cs|}/{\delta}\right)}
	{\epsilon^2} \,.$$
Then, for all $\x,\x'$, with probability of at least $1-\delta$,
$$|k_\w(\x,\x')-k_\cs(\x,\x')|\le \epsilon\,.$$
\end{theorem}
\noindent We note that if we fix the activation and assume that the depth of $\cs$ is
logarithmic, then the required bound on $r$ is polynomial. For the
ReLU activation we get a stronger bound with only quadratic dependence on
the depth. However, it requires that $\epsilon\le{1}/{\depth(\cs)}$.
\begin{theorem}\label{thm:main_ker_ReLU}
Let $\cs$ be a skeleton with ReLU activations. Let $\w$ be a random
initialization of $\cn(\cs,r)$ with
$$
	r \gtrsim \frac{\depth^2(\cs) \,
		\log\left({|\cs|}/{\delta}\right)}
	{\epsilon^2} \,.
$$
Then, for all $\x,\x'$ and $\epsilon\lesssim{}1/{\depth(\cs)}$, with
probability of at least $1-\delta$,
$$ |\kappa_\w(\x,\x')-\kappa_\cs(\x,\x')|\le \epsilon \,.$$
\end{theorem}
\noindent For the remaining theorems, we fix a $L$-Lipschitz loss
$\ell:\reals\times\cy\to [0,\infty)$.  For a distribution $\cd$ on
$\cx\times\cy$ we denote by $\|\cd\|_0$ the cardinality of the support of
the distribution. We note that $\log\left(\|\cd\|_0\right)$ is bounded by,
for instance, the number of bits used to represent an element in
$\cx\times\cy$. We use the following notion of approximation.
\begin{definition}
Let $\cd$ be a distribution on $\cx\times\cy$. A space
$\ch_1\subset\reals^\cx$ {\em $\epsilon$-approximates} the space
$\ch_2\subset\reals^\cx$ w.r.t.\ $\cd$ if for every $h_2\in\ch_2$ there is
$h_1\in\ch_1$ such that $\cl_\cd(h_1)\le \cl_\cd(h_2)+\epsilon$.
\end{definition}
\begin{theorem}\label{thm:main_dist}
Let $\cs$ be a skeleton with $C$-bounded activations. Let $\w$ be a random
initialization of $\cn(\cs,r)$ with
$$r \gtrsim \frac
	{L^4 \, R^4 \, (4C^4)^{\depth(\cs)+1}
		\log\left(\frac{LRC |\cs|}{\epsilon\delta}\right)}
	{\epsilon^4} \,.$$
Then, with probability of at least $1-\delta$ over the choices
of $\w$ we have that $\ch_\w^{\sqrt{2}R}$ $\epsilon$-approximates
$\ch^R_\cs$ and $\ch_\cs^{\sqrt{2}R}$ $\epsilon$-approximates $\ch^R_\w$.
\end{theorem}
\begin{theorem}\label{thm:main_dist_ReLU}
Let $\cs$ be a skeleton with ReLU activations and
$\epsilon\lesssim {1}/{\depth(\cc)}$. Let $\w$ be a random initialization of
$\cn(\cs,r)$ with
$$r \gtrsim \frac{L^4 \,R^4\, \depth^2(\cs)\,
	\log\left(\frac{\|\cd\|_0 |\cs|}{\delta}\right)}{\epsilon^4} \,.
$$ Then, with probability of at least $1-\delta$ over the choices of $\w$ we
have that $\ch_\w^{\sqrt{2}R}$ $\epsilon$-approximates $\ch^R_\cs$ and
$\ch_\cs^{\sqrt{2}R}$ $\epsilon$-approximates $\ch^R_\w$.
\end{theorem}
\noindent As in Theorems \ref{thm:main_ker} and \ref{thm:main_ker_ReLU}, for a fixed
$C$-bounded activation and logarithmically deep $\cs$, the required bounds on $r$
are polynomial. Analogously, for the ReLU activation the bound is polynomial
even without restricting the depth. However, the polynomial growth in
Theorems~\ref{thm:main_dist}~and~\ref{thm:main_dist_ReLU} is rather large.
Improving the bounds, or proving their optimality, is left to future work.

\subsection{Incorporating bias terms}
Our results can be extended to incorporate bias terms. Namely, in addition to the weights we can add a bias vector $\bb = \{b_v\mid v\in V\}$ and let each neuron compute the function
$$h_{v,\w,\bb}(\x) = \sigma_v\left(\textstyle
	\sum_{u\in \IN(v)}\, w_{uv}\,h_{u,\w}(\x) + b_v\right)\,.$$
To do so, we extend the definition of random initialization and compositional kernel as follows:

\begin{definition}[Random weights with bias terms]
Let $0\le \beta\le 1$. A {\em $\beta$-biased random initialization} of a neural network $\cn$ is a
multivariate Gaussian $(\bb,\w)=((w_{uv})_{uv\in E(\cn)},(b_v)_{v\in V(\cn)})$ such that each weight
$w_{uv}$ is sampled independently from a normal distribution with mean $0$
and variance ${(1-\beta)d\delta(u)}/{\delta(\IN(v))}$ if $u$ is an input neuron and ${(1-\beta)\delta(u)}/{\left(\|\sigma_{u}\|^2\,\delta(\IN(v))\right)}$ otherwise. Finally, each bias term
$b_{v}$ is sampled independently from a normal distribution with mean $0$
and variance $\beta$.
\end{definition}

\begin{definition}[Compositional kernels with bias terms]
Let $\cs$ be a computation skeleton with normalized activations and
(a single) output node $o$, and let $0\le \beta\le 1$.
For every node $v$, inductively define a kernel
$\kappa^\beta_v:\cx\times\cx\to\reals$ as follows.
For an input node $v$ corresponding to the $i$th coordinate,
define $\kappa^\beta_{v}(\x,\y)=\inner{\x^i, \y^i}$.
For a non-input node $v$, define
$$
\kappa^\beta_v(\x,\y) =
	\hat\sigma_v\left((1-\beta)
		\frac{\sum_{u\in \IN(v)}\kappa^\beta_{u}(\x,\y)}{|\IN(v)|} + \beta\right) \,.
    $$
The final kernel $\kappa^\beta_\cs$ is $\kappa^\beta_o$, the kernel associated with
the output node $o$.
\end{definition}
Note that our original definitions correspond to $\beta=0$. With the above definitions, Theorems \ref{thm:main_ker}, \ref{thm:main_ker_ReLU}, \ref{thm:main_dist} and \ref{thm:main_dist_ReLU} extend to the case when there exist bias terms. To simplify the notation, we focus on the case when there are no biases.

\section{Mathematical background}\label{sec:math}
\paragraph*{Reproducing kernel Hilbert spaces (RKHS).}
The proofs of all the theorems we quote here are well-known and can be found
in Chapter~2~of~\citep{Saitoh88} and similar textbooks. Let $\ch$ be a Hilbert space
of functions from $\cx$ to $\reals$. We say that $\ch$ is a {\em reproducing
kernel Hilbert space}, abbreviated RKHS or kernel space, if for every $\x\in
\cx$ the linear functional $f\mapsto{}f(\x)$ is bounded. The following theorem
provides a one-to-one correspondence between kernels and kernel spaces.
\begin{theorem}\label{thm:RKHS_basic}
(i) For every kernel $\kappa$ there exists a unique kernel space
$\ch_\kappa$ such that for every $\x\in \cx$,
$\kappa(\cdot,\x) \in \ch_\kappa$ and for all $f\in \ch_\kappa,\;
f(\x) = \langle f(\cdot),\kappa(\cdot,\x)\rangle_{\ch_\kappa}$.
\, (ii) A Hilbert space $\ch\subseteq \reals^\cx$ is a kernel space if
and only if there exists a kernel $\kappa:\cx\times \cx\to \reals$
such that $\ch=\ch_\kappa$.
\end{theorem}
The following theorem describes a tight connection between embeddings
of $\cx$ into a Hilbert space and kernel spaces.
\begin{theorem}\label{thm:RKHS_embedding}
A function $\kappa:\cx\times \cx\to\reals$ is a kernel if and only if there
exists a mapping $\Phi:\cx\to \ch$ to some Hilbert space for which
$\kappa(\x,\x')=\langle \Phi(\x),\Phi(\x')\rangle_{\ch}$. In addition, the
following two properties hold,
\begin{itemize}
\item $\ch_\kappa=\{f_\bv :\bv\in  \ch\}$,
  where $f_\bv(\x)=\langle \bv,\Phi (\x)\rangle_{\ch}$.
\item For every $f\in \ch_\kappa$,
  $\|f\|_{\ch_\kappa} = \inf\{\|\bv\|_{\ch}\mid f=f_\bv\}$.
\end{itemize}
\end{theorem}

\paragraph*{Positive definite functions.} A function
$\mu:[-1,1]\to\reals$ is {\em positive definite} (PSD) if there are
non-negative numbers $b_0,b_1,\ldots$ such that
$$\sum_{i=0}^\infty b_i < \infty ~ \mbox{ and } ~
  \forall x\in [-1,1],\; \mu(x)=\sum_{i=0}^\infty b_ix^i \, .$$
The {\em norm} of $\mu$ is defined as
$\|\mu\|:=\sqrt{\sum_{i} b_i}=\sqrt{\mu(1)}$.
We say that $\mu$ is {\em normalized} if $\|\mu\|= 1$
\begin{theorem}[Schoenberg, \cite{schoenberg1942positive}]\label{thm:psd_func}
A continuous function $\mu:[-1,1]\to\reals$ is PSD if and only if for all
$d=1,2,\ldots,\infty$, the function
$\kappa:\mathbb{S}^{d-1}\times\mathbb{S}^{d-1}\to\mathbb{R}$ defined by
$\kappa(\x,\x')=\mu(\inner{\x,\x'})$ is a kernel.
\end{theorem}
\noindent The restriction to the unit sphere of many of the kernels used in
machine learning applications corresponds to positive definite functions. An
example is the Gaussian kernel,
$$\kappa(\x,\x') = \exp\left(-\frac{\|\x-\x'\|^2}{2\sigma^2}\right) \,.$$
Indeed, note that for unit vectors $\x,\x'$ we have
$$\kappa(\x,\x')
  = \exp\left(-\frac{\|\x\|^2+\|\x'\|^2-2\inner{\x,\x'}}{2\sigma^2}\right)
  = \exp\left(-\frac{1-\inner{\x,\x'}}{\sigma^2}\right) \,.$$
Another example is the Polynomial kernel
$\kappa(\x,\x')=\inner{\x,\x'}^d$.

\paragraph{Hermite polynomials.} The normalized {\em Hermite polynomials} is
the sequence $h_0,h_1,\ldots$ of orthonormal polynomials obtained by
applying the Gram-Schmidt process to the sequence $1,x,x^2,\ldots$ w.r.t.\ the
inner-product 
$\inner{f,g}=\frac{1}{\sqrt{2\pi}}\int_{-\infty}^\infty
f(x)g(x)e^{-\frac{x^2}{2}}dx$.
Recall that we define activations as square integrable functions w.r.t.\
the Gaussian measure. Thus, Hermite polynomials form an orthonormal
basis to the space of activations. In particular, each activation $\sigma$
can be uniquely described in the basis of Hermite polynomials,
\begin{equation}\label{eq:hermite_expansion}
\sigma(x) = a_0h_0(x)+a_1h_1(x)+a_2h_2(x)+\ldots ~,
\end{equation}
where the convergence holds in $\ell^2$ w.r.t.\ the Gaussian measure. This
decomposition is called the Hermite {\em expansion}. Finally, we use
the following facts (see Chapter~11~in~\cite{o2014analysis} and the relevant 
\href{https://en.wikipedia.org/wiki/Hermite_polynomials}{entry} in Wikipedia):
\begin{eqnarray}
\forall n\ge 1,\;h_{n+1}(x) & =&
  \frac{x}{\sqrt{n+1}}h_n(x) - \sqrt{\frac{n}{n+1}} h_{n-1}(x) ~,
  \label{eq:hermite_recursion} \\
\forall n\ge 1,\;h'_{n}(x) & = & \sqrt{n}h_{n-1}(x)
  \label{eq:hermite_diff} \\
  \E_{(X,Y) \sim \gaussian_\rho} \hspace{-4pt} h_m(X)h_n(Y)
  & = & \begin{cases} \rho^n & n=m\\ 0 & n\ne m\end{cases}
    ~\mbox{ where }~n,m\ge 0, \, \rho\in[-1,1] ~ ,
  \label{eq:hermite_ort} \\
h_n(0) & = &
\begin{cases}
  0,  & \mbox{if }n\mbox{ is odd} \\
  \frac{1}{\sqrt{n!}}(-1)^{\tfrac{n}{2}} (n-1)!! & \mbox{if }n\mbox{ is even}
\end{cases}
~,  \label{eq:hermite_zero_val}
\end{eqnarray}
where
$$
n!! =
\begin{cases}
	1 & n \le  0 \\
	n \cdot (n-2) \cdots 5 \cdot 3 \cdot 1 & n>0 \mbox{ odd }\\
	n \cdot (n-2) \cdots 6 \cdot 4 \cdot 2 & n>0 \mbox{ even }\\
\end{cases}
\,.
$$

\section{Compositional kernel spaces}\label{sec:comp_ker}
We now describe the details of compositional kernel spaces. Let $\cs$
be a skeleton with normalized activations and $n$ input nodes associated
with the input's coordinates. Throughout the rest of the section we study
the functions in $\ch_\cs$ and their norm. In particular, we show that
$\kappa_\cs$ is indeed a normalized kernel. Recall that $\kappa_\cs$ is
defined inductively by the equation,
\begin{equation}\label{eq:recursive_ker}
\kappa_v(\x,\x') = \hat\sigma_v
  \left(
    \frac{\sum_{u\in \IN(v)}\kappa_{u}(\x,\x')}{|\IN(v)|}
  \right)\,.
\end{equation}
The recursion \eqref{eq:recursive_ker} describes a means for generating
a kernel form another kernel. Since kernels correspond to kernel spaces,
it also prescribes an operator that produces a kernel space from other kernel
spaces. If $\ch_v$ is the space corresponding to $v$, we denote this
operator by
\begin{equation}\label{eq:recursive_space}
\ch_v=\hat{\sigma}_v\left(\frac{\oplus_{u\in\IN(v)}\ch_{u}}{|\IN(v)|}\right)\,.
\end{equation}
The reason for using the above notation becomes clear in the sequel. The space
$\ch_\cs$ is obtained by starting with the spaces $\ch_{v}$ corresponding to
the input nodes and propagating them according to the structure of $\cs$,
where at each node $v$ the operation~\eqref{eq:recursive_space} is applied.
Hence, to understand $\ch_\cs$ we need to understand this operation
as well as the spaces corresponding to input nodes. The latter spaces are rather simple: for an input node $v$
corresponding to the variable $\x^i$, we have that
$ \ch_v=\{f_{\w}\mid \forall \x,\;f_\w(\x)=\inner{\w,\x^i}\}$
and
$\|f_\w\|_{\ch_v} = \|\w\|$.
To understand \eqref{eq:recursive_space}, it is
convenient to decompose it into two
operations. The first operation, termed the {\em direct average}, is
defined through the equation $\tilde{\kappa}_v(\x,\x') =
\frac{\sum_{u\in\IN(v)}\kappa_u(\x,\x')}{|\IN(v)|}$, and the resulting kernel
space is denoted $\ch_{\tilde{v}} =
\frac{\oplus_{u\in\IN(v)}\ch_{u}}{|\IN(v)|}$. The second operation, called
the {\em extension} according to $\hat{\sigma}_v$, is defined through
$\kappa_v(\x,\x') = \hat{\sigma}_v\left(\tilde{\kappa}_v(\x,\x')\right)$.
The resulting kernel space is denoted
$\ch_{v} = \hat{\sigma}_v\left(\ch_{\tilde{v}}\right)$. We next analyze these
two operations.

\paragraph{The direct average of kernel spaces.} Let $\ch_1,\ldots,\ch_n$ be
kernel spaces with kernels
$\kappa_1,\ldots,\kappa_n:\cx\times\cx\to\mathbb{R}$. Their {\em direct
average}, denoted $\ch=\frac{\ch_1\oplus\cdots\oplus\ch_n}{n}$, is the
kernel space corresponding to the kernel
$\kappa(\x,\x')=\frac{1}{n}\sum_{i=1}^n\kappa_i(\x,\x')$.
\begin{lemma}\label{lem:direct_avg}
The function $\kappa$ is indeed a kernel. Furthermore, the following
properties hold.
\begin{enumerate}
\item \label{item:dir_avg_1}
  If $\ch_1,\ldots,\ch_n$ are normalized then so is $\ch$.
\item \label{item:dir_avg_2}
  $\ch = \left\{\frac{f_1+\ldots+f_n}{n}\mid f_i\in\ch_{i}\right\}$
\item \label{item:dir_avg_3}
    $\|f\|^2_{\ch} = \inf
     \left\{ \frac{\|f_1\|^2_{\ch_1}+\ldots+\|f_n\|^2_{\ch_n}}{n}
     \mbox{ s.t. } f=\frac{f_1+\ldots+f_n}{n},\;f_i\in\ch_i\right\}$
\end{enumerate}
\end{lemma}
\proof {\bf (outline)}
The fact that $\kappa$ is a kernel follows directly from the definition of a
kernel and the fact that an average of PSD matrices is PSD. Also, it is
straight forward to verify item \ref{item:dir_avg_1}. We now proceed to
items \ref{item:dir_avg_2} and \ref{item:dir_avg_3}. By Theorem
\ref{thm:RKHS_embedding} there are Hilbert spaces $\cg_1,\ldots,\cg_n$ and
mappings $\Phi_i:\cx\to\cg_i$ such that $\kappa_i(\x,\x')=\inner{\Phi_i(\x),
\Phi_i(\x')}_{\cg_i}$. Consider now the mapping
\[
\Psi(\x) = 
  \left(\frac{\Phi_1(\x)}{\sqrt{n}},\ldots,\frac{\Phi_n(\x)}{\sqrt{n}}\right)
  \,.
\]
It holds that $\kappa(\x,\x')=\inner{\Psi(\x),\Psi(\x')}$. Properties
\ref{item:dir_avg_2} and \ref{item:dir_avg_3} now follow directly form
Thm.~\ref{thm:RKHS_embedding} applied to $\Psi$.  \proofbox

\paragraph{The extension of a kernel space.} Let $\ch$ be a normalized kernel
space with a kernel $\kappa$. Let $\mu(x)=\sum_{i} b_i x^i$ be a
PSD function. 
As we will see shortly, a function is PSD if and only if it is a
dual of an activation function.
The {\em extension} of $\ch$ w.r.t.\ $\mu$, denoted
$\mu\left(\ch\right)$, is the kernel space corresponding to the kernel
$\kappa'(\x,\x')=\mu(\kappa(\x,\x'))$.
\begin{lemma}\label{lem:extension}
The function $\kappa'$ is indeed a kernel. Furthermore, the following
properties hold.
\begin{enumerate}
\item \label{item:ext_1}
  $\mu(\ch)$ is normalized if and only if $\mu$ is.
\item \label{item:ext_2}
  $\mu(\ch) = \overline{\mathrm{span}}
    \left\{\displaystyle \prod_{g\in A}g\mid A\subset \ch,\; b_{|A|}>0 \right\}$
		where $\overline{\mathrm{span}}({\cal A})$ is the closure of the
		span of ${\cal A}$.
\item \label{item:ext_3}
  $\|f\|_{\mu(\ch)}\le\inf \left\{\displaystyle
    \sum_{A}\frac{\prod_{g\in A}\|g\|_{\ch}}{\sqrt{b_{|A|}}}
    \mbox{ s.t. } f=\sum_{A}\prod_{g\in A}g,\;A\subset\ch\right\}$
\end{enumerate}
\end{lemma}
\proof {\bf (outline)}
Let $\Phi:\cx \to \cg$ be a mapping from $\cx$ to the unit ball of a Hilbert
space $\cg$ such that $\kappa(\x,\x')=\inner{\Phi(\x),\Phi(\x')}$. Define
\[
\Psi(\x)=\left(\sqrt{b_0},\sqrt{b_1}\Phi(\x),\sqrt{b_2}\Phi(\x)\otimes \Phi(\x), \sqrt{b_3}\Phi(\x)\otimes \Phi(\x)\otimes \Phi(\x),\ldots \right)
\]
It is not difficult to verify that
$\inner{\Psi(\x),\Psi(\x')}=\mu(\kappa(\x,\x'))$. Hence, by
Thm.~\ref{thm:RKHS_embedding}, $\kappa'$ is indeed a kernel. Verifying
property \ref{item:ext_1} is a straightforward task. Properties
\ref{item:ext_2} and \ref{item:ext_3} follow by applying
Thm.~\ref{thm:RKHS_embedding} on the mapping $\Psi$. \proofbox

\section{The dual activation function} \label{dualact:sec}
The following lemma describes a few basic properties of the dual activation. These properties follow easily from the definition of the dual activation and equations
\eqref{eq:hermite_expansion}, \eqref{eq:hermite_diff}, and
\eqref{eq:hermite_ort}.
\begin{lemma}\label{lem:dual_activation}
The following properties of the mapping $\sigma\mapsto \hat\sigma$ hold:
\begin{enumerate}[label=(\alph*)]
\item If $\sigma =\sum_{i} a_i h_i$ is the Hermite expansion of
  $\sigma$, then  $\hat\sigma(\rho) = \sum_i a_i^2 \rho^i$.
	\label{lem:da_1}
\item For every $\sigma$, $\hat\sigma$ is positive definite.
	\label{lem:da_2}
\item Every positive definite function is a dual of some activation.
	\label{lem:da_3}
\item The mapping $\sigma\mapsto\hat\sigma$ preserves norms.
	\label{lem:da_4}
\item The mapping $\sigma\mapsto\hat\sigma$ commutes with differentiation.
	\label{lem:da_5}
\item For $a\in\reals$, $\widehat{a\sigma} = a^2\hat\sigma$.
	\label{lem:da_6}
\item For every $\sigma$, $\hat{\sigma}$ is continuous in $[-1,1]$ and smooth in $(-1,1)$.
	\label{lem:da_7}
\item For every $\sigma$, $\hat{\sigma}$ is non-decreasing and convex in $[0,1]$.
	\label{lem:da_8}
\item For every $\sigma$, the range of $\hat{\sigma}$ is $\left[-\|\sigma\|^2,\|\sigma\|^2\right]$.
\item For every $\sigma$, $\hat \sigma(0) = \left(\E_{X\sim N(0,1)}\sigma(X)\right)^2$ and $\hat{\sigma}(1)=\|\sigma\|^2$.
	\label{lem:da_9}
\end{enumerate}
\end{lemma}
\noindent
We next discuss a few examples for activations and calculate their dual
activation and kernel. Note that the dual of the exponential activation
was calculated in~\cite{mairal2014convolutional} and the duals of the step and the ReLU activations were calculated in~\cite{cho2009kernel}.
Here, our derivations are different and
may prove useful for future calculations of duals for other activations.

\paragraph*{The exponential activation.}
Consider the activation function $\sigma(x)=Ce^{ax}$ where $C>0$ is a
normalization constant such that $\|\sigma\|=1$. The actual value of $C$ is
$e^{-2a^2}$ but it will not be needed for the derivation below. From
properties~\ref{lem:da_5}~and~\ref{lem:da_6} of
Lemma~\ref{lem:dual_activation} we have that,
$$
\left(\hat{\sigma}\right)' = \widehat{\sigma'} =
	\widehat{a \sigma} = a^2 \hat{\sigma} \,.
$$
The the solution of ordinary differential equation
$\left(\hat{\sigma}\right)' =  a^2 \hat{\sigma}$ is of the form
$\hat{\sigma}(\rho) = b \exp\left(a^2 \rho\right)$. Since $\hat\sigma(1) = 1$
we have $b=e^{-a^2}$. We therefore obtain that the dual activation
function is
$$
\hat\sigma(\rho) = e^{a^2 \rho - a^2} = e^{a^2 (\rho - 1)} \,.
$$
Note that the kernel induced by $\sigma$ is the RBF kernel, restricted to the
$d$-dimensional sphere,
$$\kappa_\sigma (\x,\x') =
e^{a^2(\inner{\x,\x'}-1)} = e^{-\frac{a^2\|\x-\x'\|^2 }{2}} \,.$$

\paragraph*{The Sine activation and the Sinh kernel.} Consider the activation
$\sigma(x)=\sin(ax)$. We can write
$\sin(ax) = \frac{e^{iax} - e^{-iax}}{2i}$. We have
\begin{eqnarray*}
\hat \sigma (\rho) &=&
  \E_{(X,Y)\sim\gaussian_{\rho}}
    \left(\frac{e^{iaX} - e^{-iaX}}{2i}\right)
    \left(\frac{e^{iaY} - e^{-iaY}}{2i}\right)
\\
&=& -\frac{1}{4}\E_{(X,Y)\sim\gaussian_{\rho}}
  \left(e^{iaX} - e^{-iaX}\right)
  \left(e^{iaY} - e^{-iaY}\right)
\\
&=& -\frac{1}{4}\E_{(X,Y)\sim\gaussian_{\rho}}
  \left[ e^{ia(X+Y)}- e^{ia(X-Y)}-e^{ia(-X+Y)}+e^{ia(-X-Y)} \right]\,.
\end{eqnarray*}
Recall that the
characteristic function, $\E[e^{itX}]$, when $X$ is distributed $N(0,1)$
is $e^{-\frac{1}{2} t^2}$.
Since $X+Y$ and $-X-Y$ are normal variables with expectation $0$ and
variance of $2+2\rho$, it follows that,
$$\E_{(X,Y)\sim\gaussian_{\rho}}e^{ia(X+Y)} =
  \E_{(X,Y)\sim\gaussian_{\rho}}e^{-ia(X+Y)} =
  e^{-\frac{a^2(2+2\rho)}{2}} \,.$$
Similarly, since the variance of $X-Y$ and $Y-X$ is $2-2\rho$, we get
$$\E_{(X,Y)\sim\gaussian_{\rho}}e^{ia(X-Y)} =
  \E_{(X,Y)\sim\gaussian_{\rho}}e^{ia(-X+Y)} =
  e^{-\frac{a^2(2-2\rho)}{2}} \,.$$
We therefore obtain that
\[
\hat\sigma(\rho) =
  \frac{e^{-a^2(1-\rho)} - e^{-a^2(1+\rho)}}{2} = e^{-a^2}\sinh (a^2\rho)\,.
\]

\paragraph*{Hermite activations and polynomial kernels.} From Lemma
\ref{lem:dual_activation} it follows that the dual activation of the Hermite
polynomial $h_n$ is $\hat h_n(\rho)=\rho^n$. Hence, the corresponding kernel
is the polynomial kernel.

\paragraph*{The normalized step activation.}
Consider the activation
$$\sigma(x)=\begin{cases} \sqrt{2} & x>0\\ 0 & x  \le 0\end{cases} \,.$$
To calculate $\hat{\sigma}$ we compute the Hermite expansion of
$\sigma$. For $n\ge 0$ we let
\[
a_n =
\frac{1}{\sqrt{2\pi}}\int_{-\infty}^\infty\sigma(x)h_n(x)e^{-\frac{x^2}{2}}dx
= 
\frac{1}{\sqrt{\pi}}\int_{0}^\infty h_n(x)e^{-\frac{x^2}{2}}dx\,.
\]
Since $h_0(x)=1$, $h_1(x)=x$, and $h_2(x)=\frac{x^2-1}{\sqrt{2}}$,
we get the corresponding coefficients,
\begin{eqnarray*}
a_0 & = &\E_{X\sim\gaussian(0,1)}[\sigma(X)] \,=\,\frac{1}{\sqrt{2}} \\
a_1 & = &\E_{X\sim\gaussian(0,1)}[\sigma(X)X] \,=\,
  \frac{1}{\sqrt{2}}\E_{X\sim\gaussian(0,1)}[|X|] = \frac{1}{\sqrt{\pi}} \\
a_2 & = &\frac{1}{\sqrt{2}}\E_{X\sim\gaussian(0,1)}[\sigma(X)(X^2-1)]
  \,=\, \frac{1}{2}\E_{X\sim\gaussian(0,1)}[X^2-1] \,=\, 0 \,.
\end{eqnarray*}
For $n \ge 3$ we write $g_n(x)=h_n(x)e^{-\frac{x^2}{2}}$ and note that
\begin{eqnarray*}
g'_{n}(x) &=& \left[h'_n(x)-xh_n(x)\right]e^{-\frac{x^2}{2}}
\\
&=& \left[\sqrt{n}h_{n-1}(x)-xh_n(x)\right]e^{-\frac{x^2}{2}}
\\
&=& -\sqrt{n+1}\,h_{n+1}(x)e^{-\frac{x^2}{2}}
\\
&=& -\sqrt{n+1}\,g_{n+1}(x) \,.
\end{eqnarray*}
Here, the second equality follows from \eqref{eq:hermite_diff}
and the third form \eqref{eq:hermite_recursion}.
We therefore get
\begin{eqnarray*}
a_n &=& \frac{1}{\sqrt{\pi}}\int_{0}^\infty g_n(x)dx
\\
&=& -\frac{1}{\sqrt{n\pi}}\int_{0}^\infty g'_{n-1}(x)dx
\\
&=& \frac{1}{\sqrt{n\pi}}\left(g_{n-1}(0) - \overbrace{g_{n-1}(\infty)}^{=0}
\right)
\\
&=& \frac{1}{\sqrt{n\pi}}h_{n-1}(0)
\\
&=&\begin{cases}
\frac{(-1)^{\frac{n-1}{2}}(n-2)!!}{\sqrt{n\pi}\sqrt{(n-1)!}} =
\frac{(-1)^{\frac{n-1}{2}}(n-2)!!}{\sqrt{\pi n!}}& \text{if }n\text{ is odd}
\\
0 & \text{if }n\text{ is even}
\end{cases} \,.
\end{eqnarray*}
The second equality follows from \eqref{eq:hermite_recursion} and
the last equality follows from \eqref{eq:hermite_zero_val}.
Finally, from Lemma~\ref{lem:dual_activation} we have that
$\hat\sigma(\rho)=\sum_{n=0}^\infty b_n\rho^n$ where
\[
b_n=\begin{cases}
\frac{((n-2)!!)^2}{\pi n!} & \text{if }n\text{ is odd}
\\
\frac{1}{2} & \text{if }n = 0
\\
0 & \text{if }n\text{ is even }\ge 2
\end{cases} \,.
\]
In particular, $(b_0,b_1,b_2,b_3,b_4,b_5,b_6) =
\left(\frac{1}{2},\frac{1}{\pi},0,\frac{1}{6\pi},0,\frac{3}{40\pi},0\right)$.
Note that from the Taylor expansion of $\cos^{-1}$ it follows
that $\hat\sigma(\rho)= 1 - \frac{\cos^{-1}(\rho)}{\pi}$.

\paragraph*{The normalized ReLU activation.}
Consider the activation $\sigma(x)=\sqrt{2}\max(0,x)$. We now write
$\hat\sigma(\rho)=\sum_{i} b_i\rho^i$. The first coefficient is
$$b_0 = \left(\E_{X\sim\gaussian(0,1)}\sigma(X)\right)^2 =
\frac{1}{2}\left(\E_{X\sim\gaussian(0,1)}|X|\right)^2 = \frac{1}{\pi} \,. $$
To calculate the remaining coefficients we simply note that the derivative
of the ReLU activation is the step activation and the mapping
$\sigma\mapsto\hat\sigma$ commutes with differentiation. Hence, from the
calculation of the step activation we get,
\[
b_n=\begin{cases}
\frac{((n-3)!!)^2}{\pi n!} & \text{if }n\text{ is even}
\\
\frac{1}{2} & \text{if }n = 1
\\
0 & \text{if }n\text{ is odd }\ge 3
\end{cases} \,.
\]
In particular, $(b_0,b_1,b_2,b_3,b_4,b_5,b_6) =
\left(\frac{1}{\pi}, \frac{1}{2}, \frac{1}{2\pi}, 0,
  \frac{1}{24\pi}, 0, \frac{1}{80\pi}\right)$.
We see that the coefficients corresponding to the degrees $0$, $1$, and $2$
sum to $0.9774$. The sums up to degrees $4$ or $6$ are $0.9907$ and
$0.9947$ respectively. That is, we get an excellent approximation of less
than $1\%$ error with a dual activation of degree $4$.

\paragraph*{The collapsing tower of fully connected layers.}
To conclude this section we discuss the case of very
deep networks. The setting is taken for illustrative purposes but
it might surface when building networks with numerous fully connected
layers. Indeed, most deep architectures that we are aware of do not employ
more than five {\em consecutive} fully connected layers.

Consider a
skeleton $\cs_m$ consisting of $m$ fully connected layers, each layer
associated with the same (normalized) activation $\sigma$. We would like to
examine the form of the compositional kernel as the number of layers becomes
very large. Due to the repeated structure and activation we have
$$\kappa_{\cs_m}(\x,\y)=\alpha_m\left(\frac{\inner{\x,\y}}{n}\right)
~ \mbox{ where } ~
\alpha_m = \hat{\sigma}^m =
	\overbrace{\hat{\sigma}\circ\ldots\circ\hat{\sigma}}^{m\text{ times}} ~.
$$
Hence, the limiting properties of $\kappa_{\cs_m}$ can be understood from
the limit of $\alpha_m$. In the case that $\sigma(x)=x$ or $\sigma(x)=-x$,
$\hat{\sigma}$ is the identity function. Therefore $\alpha_m(\rho) =
\hat{\sigma}(\rho)=\rho$ for all $m$ and $\kappa_{\cs_m}$ is simply the
linear kernel. Assume now that $\sigma$ is neither the identity nor its
negation.  The following claim shows that $\alpha_m$ has a point-wise limit
corresponding to a degenerate kernel.
\begin{claim}
There exists a constant $0\le  \alpha_{\sigma} \le 1$ such that for all
$-1 < \rho < 1$, 
\[
\lim_{m\to\infty}\alpha_m(\rho)=\alpha_{\sigma}
\]
\end{claim}
\noindent
Before proving the claim, we note that for $\rho=1$, $\alpha_m(1)=1$ for all
$m$, and therefore $\lim_{m\to\infty}\alpha_m(1)=1$. For $\rho = -1$, if
$\sigma$ is anti-symmetric then $\alpha_m(-1)=-1$ for all $m$, and in
particular $\lim_{m\to\infty}\alpha_m(-1)=-1$. In any other case, our
argument can show that $\lim_{m\to\infty}\alpha_m(-1)=\alpha_{\sigma}$.
\proof
Recall that $\hat{\sigma}(\rho) = \sum_{i=0}^\infty b_i\rho^i$ where the
$b_i$'s are non-negative numbers that sum to 1. By the assumption that
$\sigma$ is not the identity or its negation, $b_1 < 1$.  We first claim
that there is a unique $\alpha_\sigma\in [0,1]$ such that
\begin{equation}\label{eq:babel_1}
\forall x \in (-1,\alpha_\sigma)\, ,\;\;
	\hat{\sigma}(\rho) > \rho
\text{ and }~~
\forall x \in (\alpha_\sigma,1)\, ,\;\;
	\alpha_\sigma< \hat{\sigma}(\rho) < \rho
\end{equation}
To prove~\eqref{eq:babel_1} it suffices to prove the following properties.
\begin{enumerate}[label=(\alph*)]
\item $\hat{\sigma}(\rho) > \rho$ for $\rho\in (-1,0)$
\item $\hat{\sigma}$ is non-decreasing and convex in $[0,1]$
\item $\hat{\sigma}(1)=1$
\item the graph of $\hat{\sigma}$ has at most a single intersection
		in $[0,1)$ with the graph of $f(\rho)=\rho$
\end{enumerate}
If the above properties hold we can take $\alpha_\sigma$ to be the
intersection point or $1$ if such a point does not exist.
We first show (a). For $\rho\in (-1,0)$ we have that
\begin{eqnarray*}
\hat{\sigma}(\rho) &=& b_0 + \sum_{i=1}^\infty b_i\rho^i
\;\ge \; b_0 - \sum_{i=1}^\infty b_i |\rho|^i
\; > \;  - \sum_{i=1}^\infty b_i |\rho|
\; \ge \;  - |\rho| \; = \; \rho ~.
\end{eqnarray*}
Here, the third inequality follows form the fact that $b_0 \ge 0$ and for
all $i$, $-b_i|\rho|^i \ge -b_i|\rho|$. Moreover since $b_1<1$, one of these
inequalities must be strict.
Properties (b) and (c) follows from Lemma \ref{lem:dual_activation}. Finally, to show (d), we note
that the second derivative of $\hat{\sigma}(\rho) - \rho$ is $
\sum_{i \geq 2} i(i-1)b_i \rho^{i-2}$ which is non-negative in $[0,1)$.
Hence, $\hat{\sigma}(\rho) - \rho$ is convex in $[0,1]$ and in particular
intersects with the $x$-axis at either $0$, $1$, $2$ or infinitely many times
in $[0,1]$. As we assume that $\hat{\sigma}$ is not the identity, we can
rule out the option of infinitely many intersections. Also, since
$\hat{\sigma}(1)=1$, we know that there is at least one intersection in
$[0,1]$. Hence, there are $1$ or $2$ intersections in $[0,1]$ and because one
of them is in $\rho=1$, we conclude that there is at most one intersection
in $[0,1)$.

Lastly, we derive the conclusion of the claim from equation (\ref{eq:babel_1}).
Fix $\rho\in (-1,1)$. Assume first that $\rho \ge \alpha_\sigma$. By
(\ref{eq:babel_1}), $\alpha_m(\rho)$ is a monotonically non-increasing
sequence that is lower bounded by $\alpha_\sigma$. Hence, it has a limit
$\alpha_\sigma \le \tau \le \rho < 1$. Now, by the continuity of
$\hat{\sigma}$ we have
\[
\hat{\sigma}(\tau) = \hat{\sigma}\left(\lim_{m\to\infty}\alpha_m(\rho)\right)
 = \lim_{m\to\infty}\hat{\sigma}(\alpha_m(\rho))
  = \lim_{m\to\infty}\alpha_{m+1}(\rho) = \tau \,.
\]
Since the only solution to $\hat{\sigma}(\rho) = \rho$ in $(-1,1)$ is
$\alpha_\sigma$, we must have $\tau = \alpha_\sigma$. We next deal with the
case that $-1<\rho < \alpha_\sigma$. If for some $m$, $\alpha_m(\rho)\in
[\alpha_\sigma,1)$, the argument for $\alpha_\sigma\le \rho$ shows that
$\alpha_\sigma = \lim_{m\to\infty}\alpha_m(\rho)$. If this is not the case,
we have that for all $m$, $\alpha_m(\rho)\le \alpha_{m+1}(\rho)\le
\alpha_\sigma$. As in the case of $\rho\ge \alpha_\sigma$, this can be used
to show that $\alpha_m(\rho)$ converges to $\alpha_\sigma$.
\proofbox

\section{Proofs}

\subsection{Well-behaved activations}
The proof of our main results applies to activations that are decent,
i.e.\ well-behaved, in a sense defined in the sequel.  We then show that
$C$-bounded activations as well as the ReLU activation are decent. We
first need to extend the definition of the dual activation and kernel to apply
to vectors in $\reals^d$, rather than just $\mathbb{S}^d$. We denote by
$\cm_+$  the collection of $2\times 2$ positive semi-define matrices and by
$\cm_{++}$ the collection of positive definite matrices.
\begin{definition}
Let $\sigma$ be an activation. Define the following,
\[
\bar\sigma:\cm_{+}^2\to\reals ~~ , ~~
\bar\sigma(\Sigma)=\E_{(X,Y)\sim\gaussian(0,\Sigma)}\sigma(X)\sigma(Y) ~~ , ~~
k_\sigma(\x,\y)=\bar\sigma\begin{pmatrix}
\|\x\|^2 & \inner{\x,\y}
\\
\inner{\x,\y} & \|\y\|^2
\end{pmatrix} \,.
\]
\end{definition}
\noindent
We underscore the following properties of the extension of a
dual activation.
\begin{enumerate}[label=(\alph*)]
\item The following equality holds,
	$$\hat\sigma(\rho)=\bar\sigma\begin{pmatrix}
		1 & \rho \\
		\rho & 1
	\end{pmatrix}$$

\item The restriction of the extended $k_\sigma$ to the sphere agrees
with the restricted definition.

\item The extended dual activation and kernel are defined for every
	activation $\sigma$ such that for all $a\ge 0$, $x\mapsto \sigma(ax)$ is
	square integrable with respect to the Gaussian measure.

\item For $\x,\y\in\reals^d$, if $\w\in\reals^d$ is a multivariate normal
	distribution with zero mean vector and identity covariance matrix,
	then
	$$k_\sigma(\x,\y)=\E_{\w}\sigma(\inner{\w,\x})\sigma(\inner{\w,\y}) \,.$$
\end{enumerate}
Denote
$$\cm^\gamma_+:=\left\{\begin{pmatrix}
\Sigma_{11} & \Sigma_{12}\\
\Sigma_{12} & \Sigma_{22}
\end{pmatrix}\in \cm_+\mid 1-\gamma\le \Sigma_{11},\Sigma_{22}
	\le 1+\gamma\right\} \,. $$
\begin{definition}
A normalized activation $\sigma$ is {\em
$(\alpha,\beta,\gamma)$-decent} for $\alpha,\beta,\gamma\ge 0$ if the
following conditions hold.
\begin{enumerate}[label=(\roman*)]
\item The dual activation $\bar\sigma$ is $\beta$-Lipschitz in
	$\cm_+^\gamma$ with respect to the $\infty$-norm.

\item If $(X_1,Y_1),\ldots,(X_r,Y_r)$ are independent samples from
	$\gaussian\left(0,\Sigma\right)$ for $\Sigma\in \cm_+^\gamma$ then
\[
\Pr\left(\left|\frac{\sum_{i=1}^r\sigma(X_i)\sigma(Y_i)}{r} -
	\bar\sigma(\Sigma)\right|\ge\epsilon\right)
	\le 2\exp\left(-\frac{r\epsilon^2}{2\alpha^2}\right) \,.
\]
\end{enumerate}
\end{definition}

\begin{lemma}[Bounded activations are decent]
	\label{lem:bounded_are_decent}
Let $\sigma:\reals\to\reals$ be a $C$-bounded normalized activation. Then,
$\sigma$ is $(C^2,2C^2,\gamma)$-decent for all $\gamma \ge 0$.
\end{lemma}
\proof
It is enough to show that the following properties hold.
\begin{enumerate}
\item The (extended) dual activation $\bar\sigma$ is $2C^2$-Lipschitz in
	$\cm_{++}$ w.r.t.\ the $\infty$-norm.
\item If $(X_1,Y_1),\ldots,(X_r,Y_r)$ are
 independent samples from $\gaussian\left(0,\Sigma\right)$ then
\[
\Pr\left(\left|\frac{\sum_{i=1}^r\sigma(X_i)\sigma(Y_i)}{r}-\bar\sigma(\Sigma)\right|\ge\epsilon\right) \le 2\exp\left(-\frac{r\epsilon^2}{2C^4}\right)
\]
\end{enumerate}
\noindent
From the boundedness of $\sigma$ it holds that $|\sigma(X)\sigma(Y)| \leq
C^2$. Hence, the second property follows directly from Hoeffding's bound.
We next prove the first part. Let $\z=(x,y)$ and
$\phi(\z) = \sigma(x)\sigma(y)$. Note that for
$\Sigma\in\cm_{++}$ we have
$$\bar\sigma(\Sigma) =
	\frac{1}{2\pi\sqrt{\det(\Sigma)}}
	\int_{\reals^2}\phi(\z)e^{-\frac{\z^\top\Sigma^{-1}\z}{2}}d\z \,.$$
Thus we get that,
\begin{eqnarray*}
\frac{\partial \bar\sigma}{\partial \Sigma} &=&
	\frac{1}{2\pi}\int_{\reals^2}
		\phi(\z)\left[
			\frac{\frac{1}{2}\sqrt{\det(\Sigma)}\Sigma^{-1} -
				\frac{1}{2}\sqrt{\det(\Sigma)}(\Sigma^{-1}\z\z^\top\Sigma^{-1})}
				{\det(\Sigma)}
						\right]
		e^{-\frac{\z^\top\Sigma^{-1}\z}{2}}d\z
\\
&=& \frac{1}{2\pi\sqrt{\det(\Sigma)}}\int_{\reals^2}\phi(\z)\frac{1}{2}\left[
\Sigma^{-1}-\Sigma^{-1}\z\z^\top\Sigma^{-1}
\right]e^{-\frac{\z^\top\Sigma^{-1}\z}{2}}d\z
\end{eqnarray*}
Let $g(\z)=e^{-\frac{\z^\top\Sigma^{-1}\z}{2}}$. Then, the first and second
order partial derivatives of $g$ are
\begin{eqnarray*}
\frac{\partial g}{\partial \z} & = &
	-\Sigma^{-1}\z e^{-\frac{\z^\top\Sigma^{-1}\z}{2}} \\
\frac{\partial^2 g}{\partial^2 \z} & = &
	\left[-\Sigma^{-1} +
		\Sigma^{-1}\z\z^\top\Sigma^{-1}\right]e^{-\frac{\z^\top\Sigma^{-1}\z}{2}} \,.
\end{eqnarray*}
We therefore obtain that,
\[
\frac{\partial \bar\sigma}{\partial \Sigma} =
	-\frac{1}{4\pi\sqrt{\det(\Sigma)}} \int_{\reals^2}
		\phi\frac{\partial^2 g}{\partial^2 \z} d\z \,.
\]
By the product rule we have
\[
\frac{\partial \bar\sigma}{\partial \Sigma} =
-\frac{1}{2\pi\sqrt{\det(\Sigma)}}\frac{1}{2}\int_{\reals^2}\frac{\partial^2
\phi}{\partial^2 \z} gd\z = -\frac{1}{2}\E_{(X,Y)\sim\gaussian(0,\Sigma)}\left[\frac{\partial^2 \phi}{\partial^2 \z}(X,Y)\right]
\]
We conclude that $\bar\sigma$ is differentiable in $\cm_{++}$ with
partial derivatives that are point-wise bounded by $\frac{C^2}{2}$. Thus,
$\bar\sigma$ is $2C^2$-Lipschitz in $\cm_+$ w.r.t.\ the $\infty$-norm. \qed

\medskip
We next show that the ReLU activation is decent.
\begin{lemma}[ReLU is decent]\label{lem:relu_is_decent}
There exists a constant $\alpha_\mathrm{ReLU}\ge 1$ such that for
$0\le \gamma\le 1$, the normalized ReLU activation
$\sigma(x)=\sqrt{2}\max(0,x)$ is
$(\alpha_\mathrm{ReLU},1+o(\gamma),\gamma)$-decent.
\end{lemma}
\proof
The measure concentration property follows from standard concentration
bounds for sub-exponential random variables (e.g.\ ~\cite{shalev2014understanding}).  It
remains to show that $\bar\sigma$ is $(1+o(\gamma))$-Lipschitz in
$\cm^\gamma_+$. We first calculate an exact expression for $\bar\sigma$.
The expression was already calculated in~\cite{cho2009kernel}, yet we give
here a derivation for completeness.
\begin{claim}\label{claim:relu_dual_ext}
The following equality holds for all $\Sigma\in\cm_{+}^2$,
$$\bar\sigma(\Sigma) = \sqrt{\Sigma_{11}\Sigma_{22}} \,
	\hat\sigma\!\left(\frac{\Sigma_{12}}{\sqrt{\Sigma_{11}\Sigma_{22}}}\right)
\,. $$
\end{claim}
\proof Let us denote
$$\tilde{\Sigma} = \begin{pmatrix}
1 & \frac{\Sigma_{12}}{\sqrt{\Sigma_{11}\Sigma_{12}}} \\
\frac{\Sigma_{12}}{\sqrt{\Sigma_{11}\Sigma_{12}}} & 1
\end{pmatrix} \,. $$
By the positive homogeneity of the ReLU activation we have
\begin{eqnarray*}
\bar\sigma\left(\Sigma\right) &=&
	\E_{(X,Y)\sim\gaussian(0,\Sigma)}\sigma(X)\sigma(Y) \\
&=& \sqrt{\Sigma_{11}\Sigma_{22}}
		\E_{(X,Y)\sim\gaussian(0,\Sigma)}
			\sigma\!\left(\frac{X}{\sqrt{\Sigma_{11}}}\right)
			\sigma\!\left(\frac{Y}{\sqrt{\Sigma_{22}}}\right) \\
&=& \sqrt{\Sigma_{11}\Sigma_{22}}
			\E_{(\tilde{X},\tilde{Y})\sim\gaussian\left(0,\tilde{\Sigma}\right)}
			\sigma\!\left(\tilde{X}\right)\sigma\!\left(\tilde{Y}\right) \\
&=& \sqrt{\Sigma_{11}\Sigma_{22}}\,
	\hat{\sigma}\!\left(\frac{\Sigma_{12}}{\sqrt{\Sigma_{11}\Sigma_{22}}}\right)
	\, .
\end{eqnarray*}
which concludes the proof. \qed

\medskip

For brevity, we henceforth drop the argument from $\bar{\sigma}(\Sigma)$ and
use the abbreviation $\bar{\sigma}$. In order to show that $\bar\sigma$ is
$(1+o(\gamma))$-Lipschitz w.r.t.\ the $\infty$-norm it is enough to show that
for every $\Sigma\in\cm_+^\gamma$ we have,
\begin{equation}\label{eq:grad_l1_bound}
\|\nabla \bar \sigma\|_1 =
	\left|\frac{\partial \bar\sigma}{\partial \Sigma_{12}}\right| +
	\left|\frac{\partial \bar\sigma}{\partial \Sigma_{11}}\right| +
	\left|\frac{\partial \bar\sigma}{\partial \Sigma_{22}}\right|\le
		1+o(\gamma) \,.
\end{equation}
First, Note that ${\partial \bar\sigma}/{\partial \Sigma_{11}}$ and
${\partial \bar\sigma}/{\partial \Sigma_{22}}$ have the same sign,
hence,
$$\|\nabla \bar \sigma\|_1 =
	\left|\frac{\partial \bar\sigma} {\partial \Sigma_{12}}\right| +
	\left|\frac{\partial \bar\sigma}{\partial \Sigma_{11}} +
				\frac{\partial \bar\sigma}{\partial \Sigma_{22}}\right| \,.$$
Next we get that,
\begin{eqnarray*}
\frac{\partial \bar\sigma}{\partial \Sigma_{11}} & = &
	\frac{1}{2}\sqrt{\frac{\Sigma_{22}}{\Sigma_{11}}}\,
	\hat\sigma\!\left(\frac{\Sigma_{12}}{\sqrt{\Sigma_{11}\Sigma_{22}}}\right) -
	\frac{1}{2}\sqrt{\frac{\Sigma_{22}}{\Sigma_{11}}}
		\frac{\Sigma_{12}}{\sqrt{\Sigma_{11}\Sigma_{22}}}\,
		\hat\sigma'\!\left(\frac{\Sigma_{12}}{\sqrt{\Sigma_{11}\Sigma_{22}}}\right)
\\
\frac{\partial \bar\sigma}{\partial \Sigma_{22}} & = &
	\frac{1}{2}\sqrt{\frac{\Sigma_{11}}{\Sigma_{22}}}\,
	\hat\sigma\!\left(\frac{\Sigma_{12}}{\sqrt{\Sigma_{11}\Sigma_{22}}}\right) -
	\frac{1}{2}\sqrt{\frac{\Sigma_{11}}{\Sigma_{22}}}
	\frac{\Sigma_{12}}{\sqrt{\Sigma_{11}\Sigma_{22}}}\,
	\hat\sigma'\!\left(\frac{\Sigma_{12}}{\sqrt{\Sigma_{11}\Sigma_{22}}}\right)
\\
\frac{\partial \bar\sigma}{\partial \Sigma_{12}} & = &
	\hat\sigma'\!\left(\frac{\Sigma_{12}}{\sqrt{\Sigma_{11}\Sigma_{22}}}\right)
	\, .
\end{eqnarray*}
We therefore get that the $1$-norm of $\nabla\bar\sigma$ is,
\[
\|\nabla \bar \sigma\|_1 =
\frac{1}{2}\frac{\Sigma_{11}+\Sigma_{22}}{\sqrt{\Sigma_{11}\Sigma_{22}}}
\left|
	\hat\sigma\!\left(\frac{\Sigma_{12}}{\sqrt{\Sigma_{11}\Sigma_{22}}}\right) -
	\frac{\Sigma_{12}}{\sqrt{\Sigma_{11}\Sigma_{22}}}\,
	\hat\sigma'\!\left(\frac{\Sigma_{12}}{\sqrt{\Sigma_{11}\Sigma_{22}}}\right)
\right| +
	\hat\sigma'\!\left(\frac{\Sigma_{12}}{\sqrt{\Sigma_{11}\Sigma_{22}}}\right)
	\,.
\]
The gradient of
$\frac{1}{2}\frac{\Sigma_{11}+\Sigma_{22}}{\sqrt{\Sigma_{11}\Sigma_{22}}}$
at $(\Sigma_{11},\Sigma_{22})=(1,1)$ is $(0,0)$. Therefore, from the mean
value theorem we get,
$\frac{1}{2}\frac{\Sigma_{11}+\Sigma_{22}}{\sqrt{\Sigma_{11}\Sigma_{22}}} =
	1+o(\gamma)$.
Furthermore, $\hat\sigma$, $\hat\sigma'$ and
$\frac{\Sigma_{12}}{\sqrt{\Sigma_{11}\Sigma_{22}}}$ are bounded by $1$ in
absolute value. Hence, we can write,
\[
\|\nabla \bar \sigma\|_1 =
\left|
\hat\sigma\!\left(\frac{\Sigma_{12}}{\sqrt{\Sigma_{11}\Sigma_{22}}}\right) -
\frac{\Sigma_{12}}{\sqrt{\Sigma_{11}\Sigma_{22}}}
\hat\sigma'\!\left(\frac{\Sigma_{12}}{\sqrt{\Sigma_{11}\Sigma_{22}}}\right)
\right| +
\hat\sigma'\!\left(\frac{\Sigma_{12}}{\sqrt{\Sigma_{11}\Sigma_{22}}}\right)
+ o(\gamma) \,.
\]
Finally, if we let $t=\frac{\Sigma_{12}}{\sqrt{\Sigma_{11}\Sigma_{22}}}$,
we can further simply the expression for $\nabla\bar\sigma$,
\begin{eqnarray*}
\|\nabla \bar \sigma(\Sigma)\|_1 &=& |\hat\sigma(t)-t\hat\sigma'(t)| + |\hat\sigma'(t)|  + o(\gamma)
\\
&=& \frac{\sqrt{1-t^2}}{\pi} + 1 - \frac{\cos^{-1}(t)}{\pi}  + o(\gamma) \,.
\end{eqnarray*}
Finally, the proof is obtained from the fact that the function $f(t)=\frac{\sqrt{1-t^2}}{\pi} + 1 - \frac{\cos^{-1}(t)}{\pi}$ satisfies $0\le f(t)\le 1$ for every $t\in [-1,1]$.
Indeed, it is simple to verify that $f(-1)=0$ and $f(1)=1$. Hence, it suffices to
show that $f'$ is non-negative in $[-1,1]$ which is indeed the case since,
\[
f'(t) = \frac{1}{\pi}\frac{1-t}{\sqrt{1-t^2}} =
	\frac{1}{\pi}\sqrt{\frac{1-t}{1+t}} \ge 0 \,. \qedhere
\]

\subsection{Proofs of Thms.~\ref{thm:main_ker}~and~\ref{thm:main_ker_ReLU}}
We start by an additional theorem which serves as a simple stepping stone
for proving the aforementioned main theorems.
\begin{theorem}\label{thm:ker_appr}
Let $\cs$ be a skeleton with $(\alpha,\beta,\gamma)$-decent
activations, $0<\epsilon \le \gamma$, and
$B_d = \sum_{i=0}^{d-1}\beta^i$. Let $\w$ be a random initialization of the
network $\cn=\cn(\cs,r)$ with
$$r \ge \frac{2\alpha^2B_{\depth(\cs)}^2
	\log\left(\frac{8|\cs|}{\delta}\right)} {\epsilon^2} \,. $$
Then, for every $\x,\y$ with probability of at least $1-\delta$, it holds that
\[
|\kappa_\w(\x,\y)-\kappa_\cs(\x,\y)|\le \epsilon \,.
\]
\end{theorem}
\noindent
Before proving the theorem we show that together with
Lemmas~\ref{lem:bounded_are_decent}~and~\ref{lem:relu_is_decent},
Theorems~\ref{thm:main_ker}~and~\ref{thm:main_ker_ReLU} follow from
Theorem~\ref{thm:ker_appr}. We restate them as corollaries, prove them,
and then proceed to the proof of Theorem \ref{thm:ker_appr}.
\begin{corollary}
Let $\cs$ be a skeleton with $C$-bounded activations. Let $\w$ be a random
initialization of $\cn=\cn(\cs,r)$ with
$$r \ge \frac{(4C^4)^{\depth(\cs)+1}
\log\left(\frac{8|\cs|}{\delta}\right)}{\epsilon^2} \,.$$
Then, for every $\x,\y$, w.p.\ $\ge 1-\delta$,
\[
|\kappa_\w(\x,\y)-\kappa_\cs(\x,\y)|\le \epsilon\,.
\]
\end{corollary}
\proof
From Lemma~\ref{lem:bounded_are_decent}, for all $\gamma>0$, each
activation is $(C^2,2C^2,\gamma)$-decent. By Theorem
\ref{thm:ker_appr}, it suffices to show that
$$2\left(C^2\right)^2\left(\sum_{i=0}^{\depth(\cs)-1}(2C^2)^{i}\right)^2
	\le (4C^4)^{\depth(\cs)+1} \,. $$
The sum of can be bounded above by,
\[
\sum_{i=0}^{\depth(\cs)-1}\!\!\!(2C^2)^{i} =
	\frac{(2C^2)^{\depth(\cs)}-1}{2C^2-1} \le
	\frac{(2C^2)^{\depth(\cs)}}{C^2} \,.
\]
Therefore, we get that,
\[
2\left(C^2\right)^2\left(\sum_{i=0}^{\depth(\cs)-1}\!\!\!(2C^2)^{i}\right)^2
	\le \frac{2C^4(4C^4)^{\depth(\cs)}}{C^4} \le (4C^4)^{\depth(\cs)+1} \,,
\]
which concludes the proof. \qed

\begin{corollary}
Let $\cs$ be a skeleton with ReLU activations, and $\w$ a random
initialization of $\cn(\cs,r)$ with $r \ge c_1 \frac{\depth^2(\cs)
\log\left(\frac{8|\cs|}{\delta}\right)}{\epsilon^2}$. For all $\x,\y$ and
$\epsilon\le \min(c_2,\frac{1}{\depth(\cs)})$, w.p.\ $\ge 1-\delta$,
	\[
	|\kappa_\w(\x,\y)-\kappa_\cs(\x,\y)|\le \epsilon
	\]
	Here, $c_1,c_2>0$ are universal constants.
\end{corollary}
\proof
From Lemma \ref{lem:relu_is_decent}, each activation is
$(\alpha_{\mathrm{ReLU}},1+o(\epsilon),\epsilon)$-decent. By Theorem
\ref{thm:ker_appr}, it is enough to show that
$$\sum_{i=0}^{\depth(\cs)-1}\!\!\!(1+o(\epsilon))^{i}=O(\depth(\cs)) \,.$$
This claim follows from the fact that
$(1+o(\epsilon))^{i}\le e^{o(\epsilon)\depth(\cs)}$ as long as
$i\le \depth(\cs)$. Since we assume that
$\epsilon\le{1}/{\depth(\cs)}$, the expression is bounded by $e$
for sufficiently small $\epsilon$.
\proofbox

\medskip

\noindent We next prove Theorem \ref{thm:ker_appr}.
\proof (Theorem \ref{thm:ker_appr})
For a node $u\in\cs$ we denote by $\Psi_{u,\w}:\cx\to\reals^r$ the normalized
representation of $\cs$'s sub-skeleton rooted at $u$.
Analogously, $\kappa_{u,\w}$ denotes the empirical kernel of that network.
When $u$ is the output node of $\cs$ we still use $\Psi_{\w}$ and $\kappa_\w$
for $\Psi_{u,\w}$ and $\kappa_{u,\w}$. Given two fixed $\x,\y\in\cx$ and a node
$u\in\cs$, we denote
\[
\mathcal{K}_\w^u=
\begin{pmatrix}
\kappa_{u,\w}(\x,\x) &
\kappa_{u,\w}(\x,\y)
\\
\kappa_{u,\w}(\x,\y) &
\kappa_{u,\w}(\y,\y)
\end{pmatrix},\;\; \mathcal{K}^u = \begin{pmatrix}
\kappa_u(\x,\x) & \kappa_u(\x,\y)
\\
\kappa_u(\x,\y) & \kappa_u(\y,\y)
\end{pmatrix}
\]
\[
\mathcal{K}_\w^{\leftarrow u}=\frac{\sum_{v\in\IN(u)}\mathcal{K}^{v}_\w}{|\IN(u)|}
,\quad \mathcal{K}^{\leftarrow u}=\frac{\sum_{v\in\IN(u)}\mathcal{K}^{v}}{|\IN(u)|} \,.
\]
For a matrix $\mathcal{K}\in\cm_+$ and a function $f:\cm_+\to\reals$, we denote
\[
f^p(\mathcal{K})=\begin{pmatrix}
f\!\begin{pmatrix}
\mathcal{K}_{11} & \mathcal{K}_{11}
\\
\mathcal{K}_{11} & \mathcal{K}_{11}
\end{pmatrix} & f(\mathcal{K})
\\
f(\mathcal{K}) & f\!\begin{pmatrix}
\mathcal{K}_{22} & \mathcal{K}_{22}
\\
\mathcal{K}_{22} & \mathcal{K}_{22}
\end{pmatrix}
\end{pmatrix}
\]
Note that $\mathcal{K}^u=\bar\sigma_u^p(\mathcal{K}^{\leftarrow u})$.
We say that a node $u\in \cs$, is {\em well-initialized} if
\begin{equation}\label{eq:1}
\|\mathcal{K}_\w^u- \mathcal{K}^u\|_\infty
	\le \epsilon\frac{B_{\depth(u)}}{B_{\depth(\cs)}} \,.
\end{equation}
Here, we use the convention that $B_{0}=0$. It is enough to show that with
probability of at least $\ge 1-\delta$ all nodes are well-initialized. We first note that input nodes are well-initialized by construction since
$\mathcal{K}^u_\w=\mathcal{K}^u$. Next, we show that given that all incoming
nodes for a certain node are well-initialized, then w.h.p.\ the node is
well-initialized as well.
\begin{claim}\label{claim1}
Assume that all the nodes in $\IN(u)$ are well-initialized. Then, the node
$u$ is well-initialized with probability of at least $1-\frac{\delta}{|\cs|}$.
\end{claim}
\proof
It is easy to verify that $\mathcal{K}_\w^u$ is the empirical
covariance matrix of $r$ independent variables distributed according to
$\left(\sigma(X),\sigma(Y)\right)$ where
$(X,Y)\sim\gaussian\left(0,\mathcal{K}_\w^{\leftarrow u}\right)$.
Given the assumption that all nodes incoming to $u$ are well-initialized,
we have,
\begin{eqnarray}\label{eq:2}
\left\|\mathcal{K}_\w^{\leftarrow u}-\mathcal{K}^{\leftarrow u}\right\|_\infty
&=&
\left\|
	\frac{\sum_{v\in\IN(v)}\mathcal{K}_\w^{v}}{|\IN(v)|} -
	\frac{\sum_{v\in\IN(v)}\mathcal{K}^{v}}{|\IN(v)|}\right\|_\infty\nonumber
\\
&\le& \frac{1}{|\IN(v)|}\sum_{v\in\IN(v)}
	\left\|\mathcal{K}_\w^{v}-\mathcal{K}^{v}\right\|_\infty
\\
&\le&  \epsilon\frac{B_{\depth(u)-1}}{B_{\depth(\cs)}}\nonumber \,.
\end{eqnarray}
Further, since $\epsilon\le \gamma$ then
$\mathcal{K}^{\leftarrow u}_\w\in \cm_+^\gamma$. Using the fact
that $\sigma_u$ is
$(\alpha,\beta,\gamma)$-decent and that
$r\ge \frac{2\alpha^2 B^2_{\depth(\cs)}
	\log\left(\frac{8|\cs|}{\delta}\right)}{\epsilon^2}$,
we get that w.p.\ of at least $1- \frac{\delta}{|\cs|}$,
\begin{equation}\label{eq:3}
\left\|\mathcal{K}^u_\w -
	\bar\sigma^p_u\left(\mathcal{K}_\w^{\leftarrow u}\right)\right\|_\infty
	\le \frac{\epsilon}{B_{\depth(\cs)}} \,.
\end{equation}
Finally, using \eqref{eq:2}~and~\eqref{eq:3} along with the fact that
$\bar\sigma$ is $\beta$-Lipschitz, we have
\begin{eqnarray*}
\|\mathcal{K}_\w^{u}-\mathcal{K}^{u}\|_\infty &=&
\left\|\mathcal{K}_\w^{u} -
	\bar\sigma^p_u\left(\mathcal{K}^{\leftarrow u}\right)\right\|_\infty
\\
&\le & \left\|\mathcal{K}^u_\w -
	\bar\sigma^p_u\left(\mathcal{K}_\w^{\leftarrow u}\right)\right\|_\infty +
	\left\| \bar\sigma^p_u\left(\mathcal{K}_\w^{\leftarrow u}\right) -
	\bar\sigma^p_u\left(\mathcal{K}^{\leftarrow u}\right)\right\|_\infty
\\
&\le &   \frac{\epsilon}{B_{\depth(\cs)}} +
\beta\left\| \mathcal{K}_\w^{\leftarrow u} -
	\mathcal{K}^{\leftarrow u}\right\|_\infty
\\
&\le &  \frac{\epsilon}{B_{\depth(\cs)}} +
\beta \epsilon \frac{B_{\depth(u)-1}}{B_{\depth(\cs)}}
\;=\;  \epsilon \frac{B_{\depth(u)}}{B_{\depth(\cs)}} \,. \hspace{2cm} \qed
\end{eqnarray*}

We are now ready to conclude the proof. Let $u_1,\ldots,u_{|\cs|}$ be an ordered
list of the nodes in $\cs$ in accordance to their depth, starting with the
shallowest nodes, and ending with the output node. Denote by $A_q$ the event
that $u_1,\ldots, u_q$ are well-initialized. We need to show that
$\Pr(A_{|\cs|})\ge 1-\delta$. We do so using an induction on $q$ for the
inequality $\Pr(A_q)\ge 1-\frac{q\delta}{|\cs|}$. Indeed, for $q=1,\ldots,n$,
$u_q$ is an input node and $\Pr(A_q)=1$. Thus, the base of the induction
hypothesis holds. Assume that $q>n$. By Claim (\ref{claim1}) we have that
$\Pr(A_q|A_{q-1})\ge 1-\frac{\delta}{|\cs|}$. Finally, from the induction
hypothesis we have,
\[
\Pr(A_q) \geq \Pr(A_q|A_{q-1})\Pr(A_{q-1}) \ge
	\left(1-\frac{\delta}{|\cs|}\right)
	\left(1-\frac{(q-1)\delta}{|\cs|}\right) \ge
	1-\frac{q\delta}{|\cs|} \,. \qed
\]

\subsection{Proofs of Thms.~\ref{thm:main_dist}~and~\ref{thm:main_dist_ReLU}}
Theorems~\ref{thm:main_dist}~and~\ref{thm:main_dist_ReLU} follow from using
the following lemma combined with
Theorems~\ref{thm:main_ker}~and~\ref{thm:main_ker_ReLU}. When we apply
the lemma, we always focus on the special case where one of the kernels is
constant w.p.\ $1$.
\begin{lemma} 
	\label{lem:app_ker_app_act}
Let $\cd$ be a distribution on $\cx\times\cy$, $\ell:\reals\times\cy\to\reals$
be an $L$-Lipschitz loss, $\delta>0$, and $\kappa_1,\kappa_2:\cx\times\cx\to\reals$ be two
independent random kernels sample from arbitrary distributions.
Assume that the following properties hold.
	\begin{itemize}
		\item For some $C>0$, $\forall \x\in \cx,\;
			\kappa_1(\x,\x),\kappa_2(\x,\x)\le C$.
		\item $\forall \x,\y\in
			\cx,\;\Pr_{\kappa_1,\kappa_2}
				\left(|\kappa_1(\x,\y)-\kappa_2(\x,\y)|\ge \epsilon\right)\le \tilde{\delta}$
				for $\tilde{\delta} < c_2 \frac{\epsilon^2\delta}{C^2\log^2\left(\frac{1}{\delta}\right)}$ where $c_2>0$ is a
				universal constant.
	\end{itemize}
	Then, w.p.\ $\ge 1-\delta$ over the choices of $\kappa_1,\kappa_2$, for every
	$f_1\in \ch^{M}_{\kappa_1}$ there is $f_2\in\ch^{\sqrt{2}M}_{\kappa_2}$ such that $\cl_{\cd}(f_2)\le \cl_{\cd}(f_1) + \sqrt{\epsilon}4LM$.
\end{lemma}
\noindent
To prove the above lemma, we state another lemma below followed by a basic
measure concentration result.
\begin{lemma}\label{lem:small_norm_small_l1}
Let $\x_1,\ldots,\x_m\in \reals^d$, $\w^*\in\reals^d$ and
$\epsilon>0$. There are weights $\alpha_1,\ldots,\alpha_m$
such that for $\w:=\sum_{i=1}^m\alpha_i\x_i$ we have,
\begin{itemize}
	\item $\cl(\w):=\frac{1}{m}\sum_{i=1}^m|\inner{\w,\x_i}-\inner{\w^*,\x_i}|\le\epsilon$
	\item $\sum_i |\alpha_i| \le \frac{\|\w^*\|^2}{\epsilon}$
	\item $\|\w\| \le \|\w^*\|$
\end{itemize}
\end{lemma}
\proof
Denote $M=\|\w^*\|$, $C = \max_i \|\x_i\|$, and
$y_i=\inner{\w^*,\x_i}$. Suppose that we run stochastic gradient decent on
the sample $\{(\x_1,y_1),\ldots,(\x_m,y_m)\}$ w.r.t.\ the loss $\cl(\w)$, with
learning rate $\eta = \frac{\epsilon}{C^2}$, and with projections onto the
ball of radius $M$. Namely, we start with $\w_0=0$ and at each iteration
$t\ge 1$, we choose at random $i_t\in [m]$ and perform the update,
\[
\tilde{\w}_t = \begin{cases}
\w_{t-1}-\eta\x_{i_t} & \inner{\w_{t-1},\x_{i_t}} \ge y_{i_t}
\\
\w_{t-1}+\eta\x_{i_t} & \inner{\w_{t-1},\x_{i_t}} < y_{i_t}
\end{cases}
\]
\[
\w_t = \begin{cases}
\tilde{\w}_{t} & \|\tilde{\w}_{t}\|\le M
\\
\frac{M \tilde{\w}_{t}}{\|\tilde{\w}_{t}\|} & \|\tilde{\w}_{t}\| > M
\end{cases}
\]
After $T=\frac{M^2C^2}{\epsilon^2}$ iterations the loss in expectation would
be at most $\epsilon$ (see for instance Chapter 14 in
\cite{shalev2014understanding}). In particular, there exists a sequence of
at most $\frac{M^2C^2}{\epsilon^2}$ gradient steps that attains a solution
$\w$ with $\cl(\w)\le \epsilon$.  Each update adds or subtracts
$\frac{\epsilon}{C^2}\x_i$ from the current solution. Hence $\w$ can be
written as a weighted sum of $\x_i$'s where the sum of each coefficient
is at most $T\frac{\epsilon}{C^2}=\frac{M^2}{\epsilon}$.
\proofbox

\begin{theorem}[\citet{BartlettMe02}] \label{thm:radamacher}
Let $\cd$ be a distribution over $\cx\times\cy$, $\ell:\reals\times\cy\to\reals$
a $1$-Lipschitz loss, $\kappa:\cx\times\cx\to \reals$ a kernel, and
$\epsilon,\delta>0$. Let $S=\{(\x_1,y_1),\ldots,(\x_m,y_m)\}$ be i.i.d.\
samples from $\cd$ such that
$m \ge c
	\frac{M^2 \max_{\x\in\cx}\kappa(\x,\x)+\log\left(\frac{1}{\delta}\right)}
	{\epsilon^2}$ where $c$ is a constant.
Then, with probability of at least $1-\delta$ we have,
\[
\forall f\in \ch^M_{\kappa},\; |\cl_\cd(f) - \cl_S(f)| \le \epsilon \,.
\]
\end{theorem}
\proof (of Lemma \ref{lem:app_ker_app_act})
By rescaling $\ell$, we can assume w.l.o.g that $L=1$.  Let
$\epsilon_1=\sqrt{\epsilon}M$ and $S=\{(\x_1,y_1),\ldots,(\x_m,y_m)\}\sim\cd$
be i.i.d.\ samples which are independent of the choice of
$\kappa_1,\kappa_2$. By Theorem \ref{thm:radamacher}, for a large enough
constant $c$, if $m=c \frac{C  M^2 \log\left(\frac{1}{\delta}\right)}{\epsilon_1^2}=c \frac{C\log\left(\frac{1}{\delta}\right)}{\epsilon}$,
then w.p.\ $\ge 1-\frac{\delta}{2}$ over the choice of the samples we have,
\begin{equation}\label{eq:4}
\forall f\in \ch_{\kappa_1}^{M}\cup\ch_{\kappa_2}^{\sqrt{2}M} ,\;|\cl_\cd(f) - \cl_{S}(f)|\le \epsilon_1
\end{equation}
Now, if we choose $c_2=\frac{1}{2c^2}$ then w.p.\ $\ge 1-m^2\tilde{\delta} \ge 1-\frac{\delta}{2}$
(over the choice of the examples and the kernel), we have that
\begin{equation}\label{eq:5}
\forall i,j\in [m], |\kappa_1(\x_i,\x_j)-\kappa_2(\x_i,\x_j)|< \epsilon \,.
\end{equation}
In particular, w.p.\ $\ge 1-\delta$ \eqref{eq:4} and \eqref{eq:5} hold and
therefore it suffices to prove the conclusion of the theorem under these
conditions. Indeed, let $\Psi_1,\Psi_2:\cx\to \ch$ be two mapping from $\cx$ to
a Hilbert space $\ch$ so that $\kappa_i(\x,\y)=\inner{\Psi_i(\x),\Psi_i(\y)}$.
Let $f_1\in\ch^M_{\kappa_1}$. By lemma \ref{lem:small_norm_small_l1} there are
$\alpha_1,\ldots,\alpha_m$ so that for the vector
$\w=\sum_{i=1}^m\alpha_1\Psi_1(\x_i)$ we have
\begin{equation}\label{eq:6}
\frac{1}{m}\sum_{i=1}^m|\inner{\w,\Psi_1(\x_i)}-f_1(\x_i)|\le
	\epsilon_1,\;\;\|\w\|\le M \,,
\end{equation}
and
\begin{equation}\label{eq:7}
\sum_{i=1}^m|\alpha_i|\le \frac{M^2}{\epsilon_1} \,.
\end{equation}
Consider the function $f_2\in \ch_2$ defined by $f_2(\x)=\sum_{i=1}^m\alpha_1\inner{\Psi_2(\x_i),\Psi_2(\x)}$. We note that
\begin{eqnarray*}
\|f_2\|^2_{\ch_{k_2}} &\le & \left\|\sum_{i=1}^m\alpha_i\Psi_2(\x_i)\right\|^2
\\
&=& \sum_{i,j=1}^m \alpha_i\alpha_j\kappa_2(\x_i,\x_j)
\\
&\le& \sum_{i,j=1}^m \alpha_i\alpha_j\kappa_1(\x_i,\x_j)+\epsilon\sum_{i,j=1}^m |\alpha_i\alpha_j|
\\
&=& \|\w\|^2+\epsilon \left(\sum_{i=1}^m |\alpha_i|\right)^2
\\
&\le& M^2+\epsilon \frac{M^4}{\epsilon^2_1} = 2M^2 \,.
\end{eqnarray*}
Denote by $\tilde f_1(\x) = \inner{\w,\Psi_1(\x)}$ and note that for every
$i\in [m]$ we have,
\begin{eqnarray*}
|\tilde f_1(\x_i)-f_2(\x_i)| &=& \left|\sum_{j=1}^m\alpha_j\left(\kappa_1(\x_i,\x_j)-\kappa_2(\x_i,\x_j)\right)\right|
\\
&\le &\epsilon\sum_{i=1}^m|\alpha_i|\le
	\epsilon \frac{M^2}{\epsilon_1} = \epsilon_1 \,.
\end{eqnarray*}
Finally, we get that,
\begin{eqnarray*}
\cl_{\cd}(f_2) &\le& \cl_{S}(f_2) + \epsilon_1
\\
&=& \frac{1}{m}\sum_{i=1}^m \ell\left(f_2(\x_i),y_i\right) + \epsilon_1
\\
&\le& \frac{1}{m}\sum_{i=1}^m \ell\left(\tilde{f}_1(\x_i),y_i\right) + \epsilon_1 + \epsilon_1
\\
&\le& \frac{1}{m}\sum_{i=1}^m \ell\left(f_1(\x_i),y_i\right) + |\tilde f_1(\x_i)-f_1(\x_i)| + 2\epsilon_1
\\
&\le& \frac{1}{m}\sum_{i=1}^m \ell\left(f_1(\x_i),y_i\right) + 3\epsilon_1
\\
&\le& \cl_S(f_1) + 3\epsilon_1 \le \cl_{\cd}(f_1) + 4\epsilon_1 \,,
\end{eqnarray*}
which concludes the proof.\proofbox

\section{Discussion}\label{sec:discuss}

\paragraph{Role of initialization and training.} Our results surface the
question of the extent to which random initialization accounts for the success of
neural networks. While we mostly leave this question for future research, we
would like to point to empirical evidence supporting the important role of
initialization. First, numerous researchers and practitioners demonstrated
that random initialization, similar to the scheme we analyze, is crucial to
the success of neural network learning (see for
instance~\cite{glorot2010understanding}). This suggests that starting from
arbitrary weights is unlikely to lead to a good solution. Second,
several studies show that the contribution of optimizing the representation
layers is relatively small~\cite{saxe2011random, jarrett2009best,
pinto2009high, pinto2012evaluation, cox2011beyond}. For example, competitive
accuracy on CIFAR-10, STL-10, MNIST and MONO datasets can be achieved by
optimizing merely the last layer~\cite{mairal2014convolutional,
saxe2011random}. Furthermore, Saxe et al.~\cite{saxe2011random} show that
the performance of training the last layer is quite correlated with training
the entire network. The effectiveness of optimizing solely the last layer is
also manifested by the popularity of the random features
paradigm~\cite{rahimi2009weighted}. Finally, other studies show that the
metrics induced by the initial and fully trained representations are not
substantially different. Indeed, Giryes et al.~\cite{giryes2015deep}
demonstrated that for the MNIST and CIFAR-10 datasets the distances'
histogram of different examples barely changes when moving from the
initial to the trained representation. For the ImageNet dataset the
difference is more pronounced yet still moderate.

\paragraph{The role of architecture.} By using skeletons and compositional
kernel spaces, we can reason about functions that the network can actually
learn rather than merely express. This may explain in retrospect past
architectural choices and potentially guide future choices. Let us consider
for example the task of object recognition. It appears intuitive, and is
supported by visual processing mechanisms in mammals, that in order to
perform object recognition, the first processing stages are confined to
local receptive fields. Then, the result of the local computations are
applied to detect more complex shapes which are further combined towards a
prediction. This processing scheme is naturally expressed by convolutional
skeletons.  A two dimensional version of Example~\ref{exam:diff_struct}
demonstrates the usefulness of convolutional networks for vision and speech
applications.

The rationale we described above was pioneered by LeCun and
colleagues~\cite{lecun1998gradient}. Alas, the mere fact that a network can
express desired functions does not guarantee that it can actually learn
them.  Using for example Barron's theorem~\cite{Barron93}, one may claim
that vision-related functions are expressed by fully connected two layer
networks, but such networks are inferior to convolutional networks in
machine vision applications. Our result mitigates this gap. First, it
enables use of the original intuition behind convolutional networks in order
to design function spaces that are provably learnable. Second, as detailed
in Example~\ref{exam:diff_struct}, it also explains why convolutional
networks perform better than fully connected networks.

\paragraph{The role of other architectural choices.} In addition to the
general topology of the network, our theory can be useful for understanding
and guiding other architectural choices. We give two examples. First,
suppose that a skeleton $\cs$ has a fully connected layer with the dual
activation $\hat{\sigma}_1$, followed by an additional fully connected layer
with dual activation $\hat{\sigma}_2$. It is straightforward to verify that if these
two layers are replaced by a single layer with dual activation
$\hat{\sigma}_2\circ \hat{\sigma}_1$, the corresponding compositional kernel
space remains the same. This simple observation can be useful in
potentially saving a whole layer in the corresponding networks.

The second example is concerned with the ReLU activation, which is one of the
most common activations used in practice. Our theory suggests a
somewhat surprising explanation for its usefulness. First, the dual kernel
of the ReLU activation enables expression of non-linear functions. However, this
property holds true for many activations. Second,
Theorem~\ref{thm:main_ker_ReLU} shows that even for quite deep networks with
ReLU activations, random initialization approximates the corresponding
kernel. While we lack a proof at the time of writing, we conjecture that this property holds true for many other activations. 
What is then so special about the ReLU? Well,
an additional property of the ReLU is being {\em
positive homogeneous}, i.e.\ satisfying $\sigma(ax)=a\sigma(x)$ for all
$a\ge 0$.  This fact makes the ReLU activation robust to small perturbations
in the distribution used for initialization. Concretely, if we multiply the
variance of the random weights by a constant, the distribution of the
generated representation and the space $\ch_\w$ remain the same up to a
scaling. Note moreover that training algorithms are sensitive to the
initialization.  Our initialization is very similar to approaches used in
practice, but encompasses a small ``correction'', in the form of a
multiplication by a small constant which depends on the activation.  For
most activations, ignoring this correction, especially in deep networks,
results in a large change in the generated representation. The ReLU
activation is more robust to such changes. We note that
similar reasoning applies to the max-pooling operation.

\paragraph{Future work.} Though our formalism is fairly general, we mostly analyzed
fully connected and convolutional layers. Intriguing
questions remain, such as the analysis of max-pooling and recursive
neural network components from the dual perspective. On the algorithmic side,
it is yet to be seen whether our framework can help in understanding
procedures such as dropout~\cite{srivastava2014dropout} and
batch-normalization~\cite{ioffe2015batch}. Beside studying existing
elements of neural network learning, it would be interesting to devise new
architectural components inspired by duality.
More concrete questions are concerned with quantitative improvements of the
main results. In particular, it remains open whether the dependence on
$2^{O\left(\depth (\cs)\right)}$ can be made polynomial and the quartic
dependence on $1/\epsilon$, $R$, and $L$ can be improved.  In addition to
being interesting in their own right, improving the bounds may further
underscore the effectiveness of  random initialization as a way of
generating low dimensional embeddings of compositional kernel spaces. Randomly
generating such embeddings can be also considered on its own, and we are
currently working on design and analysis of random features a la
Rahimi and Recht~\cite{RahimiRe07}.

\subsection*{Acknowledgments}
We would like to thank
Yossi Arjevani,
Elad Eban,
Moritz Hardt,
Elad Hazan,
Percy Liang,
Nati Linial,
Ben Recht, and
Shai Shalev-Shwartz
for fruitful discussions, comments, and suggestions.

\bibliographystyle{plainnat}
\bibliography{bib}

\end{document}